\documentclass[12pt]{article}
\usepackage[utf8]{inputenc}
\usepackage{authblk}

\usepackage{siunitx}
\usepackage{graphicx}
\usepackage{caption}
\usepackage{subcaption}
\usepackage{amsmath}
\usepackage{amsfonts}
\usepackage{float}
\usepackage{tabularx}
\usepackage{tablefootnote}
\usepackage[flushleft]{threeparttable}
\usepackage{multirow}
\usepackage{bigstrut}
\usepackage{url}
\usepackage{colortbl}

\usepackage{setspace}
\usepackage[margin=1in]{geometry}
\usepackage[numbers,comma,sort&compress,super]{natbib} 

\makeatletter 
\renewcommand\@biblabel[1]{#1.} 
\makeatother

\usepackage{array}
\usepackage{ragged2e}

\begin{document}
	\title{Nonlinear Manifold Learning Determines Microgel Size from Raman Spectroscopy}
	\author[1]{Eleni D. Koronaki}
	\author[2]{Luise F. Kaven}
	\author[2]{Johannes M. M. Faust}
	\author[3]{Ioannis G. Kevrekidis} 
	\author[,4,2,5]{Alexander Mitsos*}
	
	\affil[1]{Faculté des Sciences, de la Technologie et de la Communication, Université de Luxembourg, Maison du Nombre, Avenue de la Fonte 6, L-4364 Esch-sur-Alzette, Luxembourg}
	\affil[2]{Process Systems Engineering (AVT.SVT), RWTH Aachen University, Forckenbeckstr. 51, 52074 Aachen, Germany}
	\affil[3]{Department of Chemical and Biomolecular Engineering and Department of Applied Mathematics
		and Statistics, Whiting School of Engineering, Johns Hopkins University, 3400 North Charles Street,
		Baltimore, MD 21218, USA.}
	\affil[4]{JARA-CSD, 52056 Aachen, Germany}
	\affil[5]{Institute of Energy and Climate Research, Energy Systems Engineering (IEK-10), Forschungszentrum Jülich GmbH, 52425 Jülich, Germany}
	
	\date{\today}
	
	\maketitle
	
	\section*{Abstract}
	
	Polymer particle size constitutes a crucial characteristic of product quality in polymerization.
	Raman spectroscopy is an established and reliable process analytical technology for in-line concentration monitoring.
	Recent approaches and some theoretical considerations show a correlation between Raman signals and particle sizes but do not determine polymer size from Raman spectroscopic measurements accurately and reliably.
	With this in mind, we propose three alternative machine learning workflows to perform this task, all involving diffusion maps, a nonlinear manifold learning technique for dimensionality reduction: $(i)$ directly from diffusion maps, $(ii)$ alternating diffusion maps, and $(iii)$ conformal autoencoder neural networks.
	We apply the workflows to a data set of Raman spectra with associated size measured via dynamic light scattering of 47 microgel (cross-linked polymer) samples in a diameter range of \SIrange[]{208}{483}{\nano\meter}.
	The conformal autoencoders substantially outperform state-of-the-art methods and results for the first time in a promising prediction of polymer size from Raman spectra.

	\section*{Introduction}
	
	Process analytical methods are crucial for optimizing process performance and product properties, especially in polymerization.
	In-line spectroscopic methods are advantageous, and, in particular, near-infrared~(NIR) and Raman spectroscopy are widely applied spectroscopic methods, see, e.g., the reviews.~\citep{Chew.2010, Beer.2011, Pomerantsev.2012, Simon.2015}
	Evaluation methods for concentrations from (either on-line or off-line) spectroscopic data are established and comprise regression models, such as univariate peak integration based on the Beer-Lambert law,~\citep{Beer.1852} multivariate partial least squares~(PLS), or artificial neural networks~(ANNs),~\citep{Marini.2008} and physically supported strategies such as multivariate curve resolution-alternating least squares~(MCR-ALS)~\citep{Garrido.2008} or indirect hard modeling~(IHM).~\citep{Alsmeyer2004} 
	
	Size is a crucial product feature in several processes, e.g., polymerization and crystallization. 
	In contrast to concentrations, the size prediction from spectroscopic data remains a major challenge.
	Herein, we consider cross-linked polymer networks called microgels, which are not considered particles.~\citep{Keidel.2018}
	Thus, we use the term \textit{microgel size} or \textit{polymer size} instead of the more common \textit{particle size}.
	The fact that particles such as polymers influence spectroscopic measurements through light scattering is well established,~\citep{Bohren.1998} and experimental evidence of the correlation between Raman scattering and polymer size has been presented,~\citep{Reis.2003} even for relatively large particles.~\citep{vandenBrink.2002}
	However, only a few approaches attempt to predict polymer sizes from Raman spectra.~\citep{Reis.2003, vandenBrink.2002, Ito.2002, Houben.2015, Ambrogi.2017, MeyerKirschner.2018}
	These approaches rely on relatively small sets of data points and focus on training data-driven models.
	As most literature studies do not provide performance metrices or alternatively raw data publication, a comprehensive comparison of the quantitative performance metrices of studies from the literature is not feasible.
	However, qualitatively a lack of prediction accuracy of these approaches is observed compared to established methods such as dynamic light scattering~(DLS).
	An overview of the state-of-the-art work on polymer size prediction from Raman spectra is presented in Tab.~\ref{tab:StateOfArt}.
	Most of these methods are based on the linear methods PLS or principal component analysis~(PCA), which reduce the predictors to a smaller set of uncorrelated components and perform least squares regression on these components instead of on the original data.
	

	Each spectral measurement consists of many measured intensities, resulting in a large dimensionality of the input vector.
	However, the measured intensities are not independent and ideally depend on a small number of meaningful properties, e.g., concentration and size.
	When the available data live in a low-dimensional, yet nonlinear manifold, linear methods often fail to capture the majority of the variance of the data, even with an increased number of principal components.
	Hence, we propose using nonlinear manifold learning approaches to achieve significant dimensionality reduction and, more importantly, to identify latent variables possessing specific desired properties.
	Dimensionality reduction replaces extensive data sets with a handful of latent variables.
	For the reduction, we employ diffusion maps~(DMAPs),~\citep{r19,r20,r21} to derive a parsimonious reduced description of the Raman spectra. 
	Then, as one alternative, we predict the polymer size directly from the latent variables computed with the DMAPs algorithm.
	Moreover, we take additional steps regarding the characteristics of the latent space by applying two alternative machine-learning methods to discover common latent variables between the spectra and the polymer sizes. 
	In this sense, \textit{common} describes a set of variables that has a one-to-one correlation to the desired observed quantity. 
	The common variable between the spectra and the polymer size is used as a data-driven junction through which we can predict the polymer size given the spectrum of a new sample.
	
\begin{figure}[ht!]
		\centering
		\includegraphics[width=0.6\textwidth]{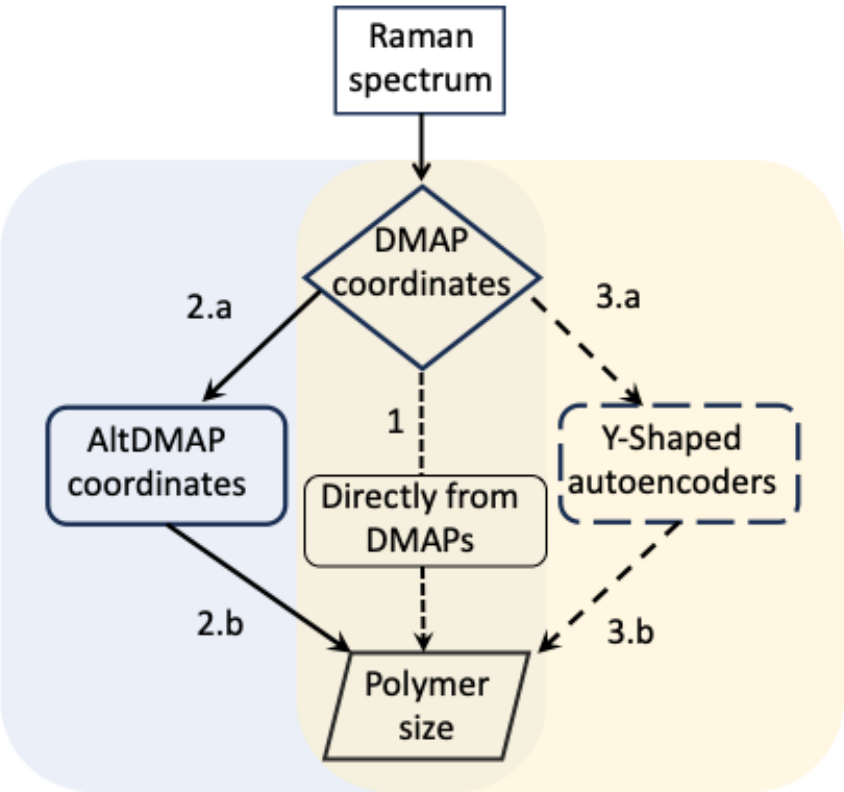}
		\caption{Schematic overview of the proposed machine learning approaches applied to low dimensional parameterization (provided by DMAPs) of measured Raman spectra: (i) in Approach 1 the polymer size is predicted directly from DMAPs, (ii) Approach 2 implements AltDMAPs to find the data-driven common variable that is one-to-one with the polymer size (2.a), and in (2.b) predicts the polymer size from this common variable, and (iii) Approach 3 implements a Y-shaped conformal autoencoder, that identifies a ``designer" latent space (3.a) from which it is possible to predict the polymer size (3.b).}
		\label{fig:flowchartMethods}
	\end{figure}

	We propose three alternative machine learning methods (presented schematically in Fig.~\ref{fig:flowchartMethods}): ($i$) direct prediction from DMAP coordinates;
	($ii$) an alternating diffusion maps~(AltDMAPs) algorithm, initially introduced by Lederman et al.,~\citep{lederman2018learning} and later in the context of multimodal data fusion~\citep{katz2019alternating} und identification of ``jointly smooth" functions on input and output manifolds;~\citep{dietrich2022spectral}
	and ($iii$) an ensemble of concurrently trained neural networks (NNs), named Y-shaped conformal autoencoders, introduced by Evangelou et al.~\citep{yshaped2022} in the context of parameter non-identifiability. 
	
	Methods ($ii$) and ($iii$) are particularly appropriate for measurements that depend on various combinations of factors, the effect of which is not readily quantifiable. 
	Herein, these methods are employed to identify the changes in the spectra explicitly attributed to the polymer sizes since the samples are not pretreated.
	As several factors beyond the size influence, e.g., concentrations of monomer, inhibitor, surfactant, and other partaking species in the reaction, influence the spectra, the proposed methods act as a nonlinear filter that isolates (in a sense, disentangles) the spectral changes that are attributed to the differences in polymer size.
	
	The common starting point for the machine learning approaches proposed in this work addresses the reduction of the high dimensionality of spectra, here achieved with DMAPs.
	The DMAPs algorithm is based upon (mathematical) diffusion processes on the data and facilitates discovering meaningful low-dimensional intrinsic geometric descriptions of data sets, even when the data is high-dimensional, nonlinear, and corrupted by (relatively small) noise. 
	The algorithm has been successfully used for dimensionality reduction in applications relevant to reaction engineering in \citep{r40,KORONAKI2023108357}, among others. 
	We propose its use as an effective dimensionality reduction of full spectra, consisting of $\sim$~11,000 wavenumbers. This reduction enables efficient interpolation and regression since much fewer (typically $<$10) variables are involved.
	Once the low-dimensional representation of the spectra is determined in an \textit{offline step}, it is possible to translate between coordinates in the ambient~(spectra) and the reduced space~(DMAP coordinates).
	The mapping from high to low dimension is achieved with the Nyström extension,~\citep{nystrom1929praktische,fowlkes2001efficient} whereas for the inverse, a particular implementation of Geometric Harmonics, called DoubleDMAPs, is selected.~\citep{evangelou2022double}
	Accurate reconstruction of the data set from the selected DMAP coordinates indicates that the latter is an adequate low-dimensional parameterization.
	
	After the dimensionality reduction, three alternative machine learning workflows ($i$) to ($iii$) are compared.
	The first approach involves regression analysis to estimate the relationship between the latent variables (DMAP coordinates) and the polymer size (cf. Approach 1 denoted as ``Directly from DMAPs'' in Fig.~\ref{fig:flowchartMethods}). 
	For the implementation of AltDMAPs, a common variable between spectra (in their reduced representation) and polymer size is found here as an \textit{offline} step (cf. Step 2.a in Fig.~\ref{fig:flowchartMethods}). 
	Then, the overall \textit{online} computational workflow to predict the polymer size from a new spectrum starts by determining its low-dimensional DMAP coordinates with the Nyström extension.
	From that, the common variable, the AltDMAP coordinate, is predicted with an ANN or other regression methods, such as XGBOOST  (cf. Step 2.a in Fig.~\ref{fig:flowchartMethods}). 
	Finally, the corresponding size is predicted from the AltDMAP coordinates with an ANN or with DoubleDMAPs (cf. Step 2.b in Fig.~\ref{fig:flowchartMethods}). 
	
	In approach ($iii$), we exploit recent advances in conformal autoencoder neural network techniques.
	The DMAP coordinates are used as inputs (and also outputs) to a Y-shaped autoencoder to disentangle the dependencies of the latent variables discovered by a traditional autoencoder architecture (cf. Step 3.a in Fig.~\ref{fig:flowchartMethods}) 
	and define the desired output, i.e., the polymer size, as a function of a latent variable (cf. Step 3.b in Fig.~\ref{fig:flowchartMethods}). 
	Ultimately, given a new set of DMAPs, this ``designer'' neural network (NN) predicts the corresponding polymer size (and reconstruct the DMAPs themselves). 
	
	We compare the proposed workflows with state-of-the-art techniques and show that it is essential to not only find \textit{a generic} parsimonious low-dimensional parameterization of the data (here achieved with DMAPs) but to find the \textit{appropriate one}, possessing a component that the polymer size can be written as a function of.
	We demonstrate that although the number of pairs of microgel size and spectral measurements is moderate (47, i.e., at least as high as in previous works, Tab.~\ref{tab:StateOfArt}), the workflow is a promising direction in predicting polymer size in-line from Raman spectra. 
	
	The remainder of the article is structured as follows: First, the process of data collection is presented along with an overview of the state-of-the-art methods for polymer size prediction from spectra.
	Subsequently, the DMAPs and AltDMAPs methods are summarized, followed by a detailed description of the proposed workflow.
	The conformal autoencoder architecture is then presented, building on the successful dimensionality reduction achieved with DMAPs.
	Finally, results and conclusions are drawn from the proposed implementation. 
	
	\section*{Methods}
	
	The following sections include the description of the data set used in this contribution and the applied methods for size prediction: benchmark methods, and the proposed workflows, including the DMAPs approach, AltDMAPs workflow, and the conformal autoencoder.
	
	\subsection*{Data} \label{sec:data}
	We employ data from our samples taken from continuous synthesis of microgels.~\citep{Kaven.2023}
	The considered microgels here are based on N-isopropylacrylamide and cross-linked via \textit{N},\textit{N}'-methylene\-bis\-(acryl\-amide).
	The reactor and measurement setup for the continuous synthesis are explained in our previous work.~\citep{Kaven.2021}
	Using the continuous flow reactor, microgels of different sizes are synthesized by changing the reactor temperature and flow rates and the initiator and surfactant concentration.
	As microgels are known to be of monodisperse size,~\citep{Kather.2018,Wolff.2018,Janssen.2019} they represent an excellent system to study polymer size predictions from Raman spectroscopy.
	In our previous works,~\citep{Kaven.2023} we conducted in-line Raman spectra at reaction temperature at \SIrange{60}{80}{\degreeCelsius} and DLS measurements at \SI{50}{\degreeCelsius}.
	In contrast, in the present work, we acquire additional off-line Raman measurements of the same samples but at \SI{20}{\degreeCelsius} and restricted conditions.
	The restrictions include measuring the samples all within a small amount of time (over two experimentation days) and in glass vials filled to the exact same fluid level to ensure equal conditions for the acquisition of all spectra and to eliminate external influences on the measurements.
	Consequently, we also conduct further DLS measurements at \SI{20}{\degreeCelsius}.
	
	The data consist of Raman spectra and DLS measurements of microgel samples.
	The samples are taken from the output of the continuous flow reactor, which runs at different experimental conditions for each sample.
	The samples are measured off-line without further treatment, e.g., filtration or dialysis.
	In total, we use the data from 47 samples at different microgel sizes in the range of \SIrange[]{208}{483}{\nano\meter} in diameter.
	The determined polydispersity index of these microgels ranges well below 0.368, indicating monodisperse size distribution.
	
	Microgels have a different size depending on a threshold in temperature: above approximately \SI{32}{\degreeCelsius}, they occur in a collapsed state; below the threshold temperature, they occur in a swollen state.
	Hence, the microgel sizes at \SI{20}{\degreeCelsius} are almost twice as big as at the reaction temperature.
	Each sample is measured three times via Raman spectroscopy.
	Raman spectra are taken with an acquisition time of \SI{40}{\second} using an RXN2 Raman Analyzer (Kaiser Optical Systems, Ann Arbor, Michigan, USA) with cosmic ray correction.
	DLS measurements of the samples diluted in ultrapure water are conducted via the Zetasizer Ultra (Malvern Panalytical, Malvern, UK) at \SI{20}{\degreeCelsius} with a scattering angle of \SI{90}{\degree}.
	Each DLS measurement is repeated four times, and the software ZS Xplorer analyzes the scattering intensities.
	
	The Raman spectra comprise the Raman intensity measured over a range of wavelengths.
	The global range is between \SIrange{100}{3425}{\per\centi\meter} correlating to 11,084 wavelengths per spectrum.
	Different spectra pretreatment methods can be applied to the spectral data.
	We compare using raw spectra and spectra with a linear fit or rubber band baseline correction in combination with standardization in the form of either Standard Normal Variate (SNV) or Min-Max normalization.
	The experimental data set is published open access~\citep{dataKaven.2023} and comprises raw and pretreated Raman and evaluated DLS data.
	
	We use the same data set for all subsequently described methods to predict microgel size from Raman spectra.
	Out of the 47 pairs of microgel size and Raman spectra, we take 40 for training and 7 for testing.
	The same split is applied to quantify the prediction performance of each considered method (state-of-the-art methods and our proposed workflow with nonlinear methods).
	We conduct the training for each prediction method with 10-fold cross-validation, which involves splitting the training set into 10 smaller sub-sets and using 9 for training and 1 for testing. By repeating this process, using a different collection of sub-sets for training and validation each time, it is possible to define the best possible model hyper-parameters without sacrificing a lot of data. The number of hyper-parameters varies depending on the prediction method.
	Hence, the set of hyper-parameters for the individual method is described for each method separately in the following sections.
	The prediction performance of each method is evaluated based on commonly applied metrics: coefficients of determination~(R\textsuperscript{2}), root mean squared error~(RMSE), and mean absolute percentage error~(MAPE) for training and testing.
	
	The prediction accuracy is reflected in the $\%$-error calculated as:
	\begin{equation}
		\%\mathrm{-error} = 100 \cdot \frac{D_\mathrm{H}^\mathrm{predicted}-D_\mathrm{H}^\mathrm{actual}}{D_\mathrm{H}^\mathrm{actual}},
	\end{equation}
	where $D_\mathrm{H}$ is the microgel's hydrodynamic diameter.
	Based on the $\%$-error the MAPE is calculated as the sum of the $\%$-errors divided by the number of observations.

	\subsection*{Benchmark Methods for Polymer Size Prediction from Raman Spectra} \label{sec:benchmark}
	To benchmark our proposed method, we compare it against two state-of-the-art methods.
	These methods include the direct application of PLS to the spectral intensities and the application of PLS to fitted IHM parameters.
	
	\subsubsection*{Partial Least Squares Regression of Spectral Intensities} \label{sec:direct_PLS}
	
	As conducted in the literature,~\citep{Reis.2003,Ito.2002,Houben.2015,Ambrogi.2017} we apply a partial least squares (PLS) model regression directly to the spectral data as first introduced by Ito et al..~\citep{Ito.2002}
	We consider different spectral ranges for the calibration of our PLS models.
	These ranges include the global range and the so-called fingerprint (FP) region between \SIrange{850}{1800}{\per\centi\meter}.
	Also, we apply pretreatment methods, combining two different types of baseline subtractions (linear fit and rubber band) and two normalization approaches (MinMax and SNV).
	Further, we normalize the data using the \texttt{zscore} function in Matlab.
	We analyze the results based on the metrics R\textsuperscript{2}, RMSE, and MAPE for calibration and validation.
	Based on the MSE for cross-validation, we chose the number of components (latent variables) for the PLS regression.

	\subsubsection*{Regression of Hard Model Parameters} \label{sec:PLS_IHM}
	We conduct the regression of fitted hard model parameters for predicting microgel sizes, as we proposed in previous works.~\citep{MeyerKirschner.2018}
	First, an IHM~\citep{Alsmeyer2004} is developed.
	For that model, the spectral range of the FP is considered.
	The range between \SIrange[]{1552}{1560}{\per\centi\meter} is excluded as it is attributed to an atmospheric oxygen signal.
	Besides the range restrictions, no further pretreatment is usually applied as per our findings.~\citep{MeyerKirschner.2018}
	However, we compare the PLS performance of the IHM parameters with and without pretreatment for a comprehensive comparison.
	The applied pretreatment for the comparison is MinMax or SNV normalization and linear fit or rubber band baseline subtraction.
	The IHM prediction model is calibrated using calibration measurements from our previous work.~\citep{dataKaven.2021}
	
	The model includes component hard models for the monomer, polymer, and water and a linear baseline.
	The hard model of each component consists of multiple characteristic peaks.
	The individual peaks are characterized by four parameters: position, intensity, shape (fraction of the Gaussian part), and half-width at half maximum (HWHM).
	The complete indirect hard model combines the component models with their distinctive peaks.
	The indirect hard model parameters are adjusted to suit the spectra of interest within the fitting process.
	The applied fitting mode constitutes a medium interaction method, where the weights of the components, the baseline, and the peak positions are varied. 
	The weights of the components represent the magnitude of the individual component in the spectra.
	Each component (monomer, polymer, water) is accredited with one weight parameter during the model fitting process.
	The incorporated linear baseline is fitted with regard to its offset and slope.
	In this context, we restrict component shifts to avoid ambiguities due to overlapping spectral peak positions.
	The fitting mode follows our previous work.~\citep{MeyerKirschner.2018}
	The medium interaction fitting mode results in 49 modified parameter values (intercept and slope of the baseline, one weight for each of the three components, 23 monomer peak positions, 17 polymer peak positions, and four water peak positions) that serve as the input variables to the PLS regression model.
	
	In addition, we compare the medium fitting mode with the fitting mode with high interaction.
	For the high interaction, in addition to the changes in the medium interaction, all peak parameters can be varied within the fitting process.
	Thus, the high interaction method results in 181 modified parameter values (intercept and slope of the baseline, one weight for each of the three components, 23 monomer peaks, 17 polymer peaks, and four water peaks, where each peak is characterized by the four parameters described previously).
	
	The fitted IHM parameter values serve as input to the subsequent PLS regression.
	Again, we normalize the data using the \texttt{zscore} function in Matlab.
	We also analyze the results of the PLS regression based on the R\textsuperscript{2}, RMSE, and MAPE values for the hybrid modeling approach, combining IHM and PLS.
	Based on the MSE for cross-validation, we choose the number of components for the PLS regression.
	
	\subsection*{Diffusion Maps} \label{sec:DiffusionMaps_workflow}
	
	The following paragraphs highlight the functionality of DMAPs for dimensionality reduction.
	In addition, we establish a workflow for predicting polymer sizes directly from DMAP coordinates.

	\subsubsection*{Diffusion Maps for Dimensionality Reduction}
	\label{sec:Diffusion_Maps}
	The DMAPs framework has been shown to facilitate the discovery of meaningful low-dimensional intrinsic geometric descriptions of data sets~.\citep{KORONAKI2023108357,evangelou2022double,psarellis2022data}
	For a detailed description of the method, the interested reader is referred to the seminal papers~\citep{r19,coifman2006geometric} and also, among others,~\citep{r21, r40, KORONAKI2023108357, evangelou2022double}.
	Here, we use DMAPs to discover the dimensionality of the manifold that contains a collection of Raman spectra of microgel samples.
	Furthermore, DMAPs discover data-driven coordinates on (i.e., parameterizations of) the low-dimensional manifold where the data reside. 
	These coordinates do not necessarily have a physical meaning, i.e., they do not have to correlate to any physical quantity.
	The coordinates of the manifold are a few of the leading eigenvectors, $\phi_i$, of a scaled affinity matrix, which contains the Euclidean distances between all the pairs of available data points.
	
	Discovering which eigenvectors parameterize independent directions and do not span the same direction with different frequencies (harmonics) is essential.
	To distinguish independent eigenvectors, the local linear regression algorithm can be used as proposed by Dsilva et al.~\citep{r25}
	
	Notably, the proposed approach includes reverse mapping from the DMAP coordinates to the variables in ambient space, which allows for the translation between the high and low-dimensional data description.
	To this end, Geometric Harmonics are proposed, introduced initially in \citep{r19}, as a scheme for extending functions defined on data $\textbf{X}$, $f(\textbf{X}):\textbf{X} \to \textbf{R}$, for $x_{new} \notin \textbf{X}$.
	Here, the DoubleDMAPs~\citep{evangelou2022double} algorithm, a particular implementation of Geometric Harmonics, is selected.
	DoubleDMAPs are suitable for low-dimensional data that can be parameterized by just a few non-harmonic eigenvectors.
	Generally, Geometric Harmonics construct an input-output mapping between the ambient coordinates $\textbf{X}$ and a function of interest $f$ defined on $\textbf{X}$ by operating directly on the non-harmonic DMAP coordinates.
	
	\subsubsection*{Prediction directly from DMAPs}
	\label{sec:simpledmaps}
	Establishing a parsimonious embedding of the spectra enables polymer size prediction directly from the DMAP coordinates.
	The direct prediction based on DMAP coordinates is achieved here with two different regression methods: (a) NN and (b) XGBOOST, using the DMAP coordinates as input and the polymer size as output.
	The schematic representation of the direct workflow is shown in Fig.~\ref{fig:onlineDirecltyfromDMAPs}.
	
	This direct approach is expected to perform well when the latent variables derived by DMAPs possess a one-to-one dependence on the desired observable, here the polymer size.
	
	
	\subsection*{Alternating Diffusion Maps Workflow}\label{sec:altdmaps_workflow}
	The following sections introduce to the AltDMAPs algorithm and describe the prediction workflow based on AltDMAPs for identifying common variables between the spectra and the polymer size.
	
	\subsubsection*{Alternating Diffusion Maps} \label{sec:altDMAPs}
	AltDMAPs is a method, based on DMAPs, designed to parameterize the common variable between independent sets of observations, $\{s^{(1)}_i,s^{(2)}_i\}$. 
	The method is introduced by Lederman et al.,~\citep{lederman2018learning} and the interested reader is referred to this paper and also to \citep{TALMON2019848} for a detailed presentation of the method.
	Here, only key points are presented for completeness.
	
	At the core of the methodology lies the established DMAPs algorithm, which starts with a data set of $N$ individual points (represented as $d$-dimensional real vectors, $x_1,..., x_{N}$). 
	A similarity measure between each pair of vectors $(x_i, x_j)$, is computed using the standard Euclidean distance, based on which an affinity matrix is constructed.
	
	A popular choice is the Gaussian kernel $w(i,j)=\exp\left[-\left(\frac{||x_i-x_j||}{\epsilon}\right)^2\right]$ where $\epsilon$ is a hyper-parameter that quantifies the kernel's bandwidth.
	To recover a parameterization insensitive to the sampling density, the normalization
	\begin{equation*}
		\widetilde{\textbf{W}} = \textbf{P}^{-1} \textbf{W} \textbf{P}^{-1}
	\end{equation*}
	is performed, where $P_{ii} = \sum_{j=1}^N W_{ij}$ and $W_{ij}$ the elements of the matrix $\textbf{W}$.
	A second normalization applied on $\widetilde{\textbf{W}}$,
	\begin{equation}
		\textbf{K}=\textbf{D}^{-1} \widetilde{\textbf{W}}
		\label{eq:kernel}
	\end{equation}
	gives a \(N\times N\) Markov matrix \textbf{K}; here \(\textbf{D}\) is a diagonal matrix, collecting the row sums of the matrix \(\widetilde{\textbf{W}}\) eigenvectors $\phi_i$.
	
	Based to the DMAPs algorithm, the AltDMAPs algorithm defines two weighted kernel matrices, as defined in equation \eqref{eq:kernel}: one that is based on the measurements from $s^{(1)}$ and the other based on the measurements from $s^{(2)}$, and constructs the alternating-diffusion operator as the product of the two normalized weight matrices:
	\begin{equation*}
		\textbf{K}^{alt}=\textbf{K}^{(1)}\textbf{K}^{(2)}.
	\end{equation*}
	It has been shown that the diffusion process defined by this operator is equivalent to one that would have been created from measurements of only the common variable~\citep{lederman2018learning}.
	This finding can be further explained in terms of alternating propagation steps using $K^{(1)}$ followed by $K^{(2)}$.
	Each propagation step can be considered a Markov chain that transitions to similar samples, as prescribed by the respective similarity matrix, first in the first sample set, followed by the second.
	
	\subsubsection*{Prediction Workflow based on AltDMAPs} \label{sec:workflow}
\begin{figure}[H]
		\centering
		\begin{subfigure}[b]{\textwidth}
			\centering
			\includegraphics[width=\textwidth]{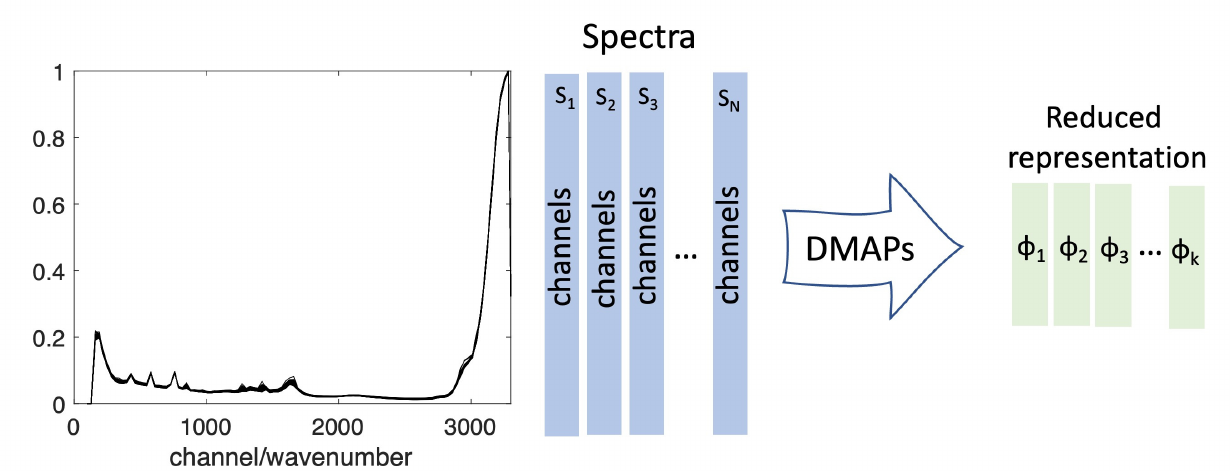}
			\caption{}
			\label{fig:off1}        
		\end{subfigure}
		\vfill
		\begin{subfigure}[b]{\textwidth}
			\centering
			\includegraphics[width=\textwidth]{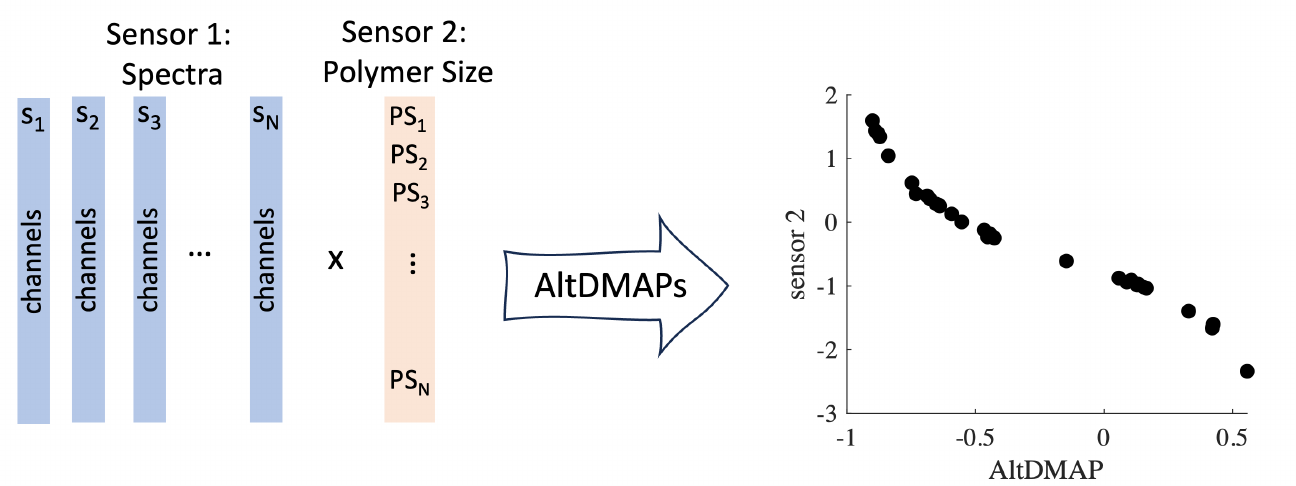}
			\caption{}
			\label{fig:off2}        
		\end{subfigure}
		\caption{Schematic of the \textit{offline} steps: (a) Step 1: Finding a reduced representation of the spectra with DMAPs; (b) Step 2: Finding the common variable between spectra and polymer size with AltDMAPs.}
		\label{fig:offline}
	\end{figure}

\begin{figure}[H]
		\centering
		\includegraphics[width=\textwidth]{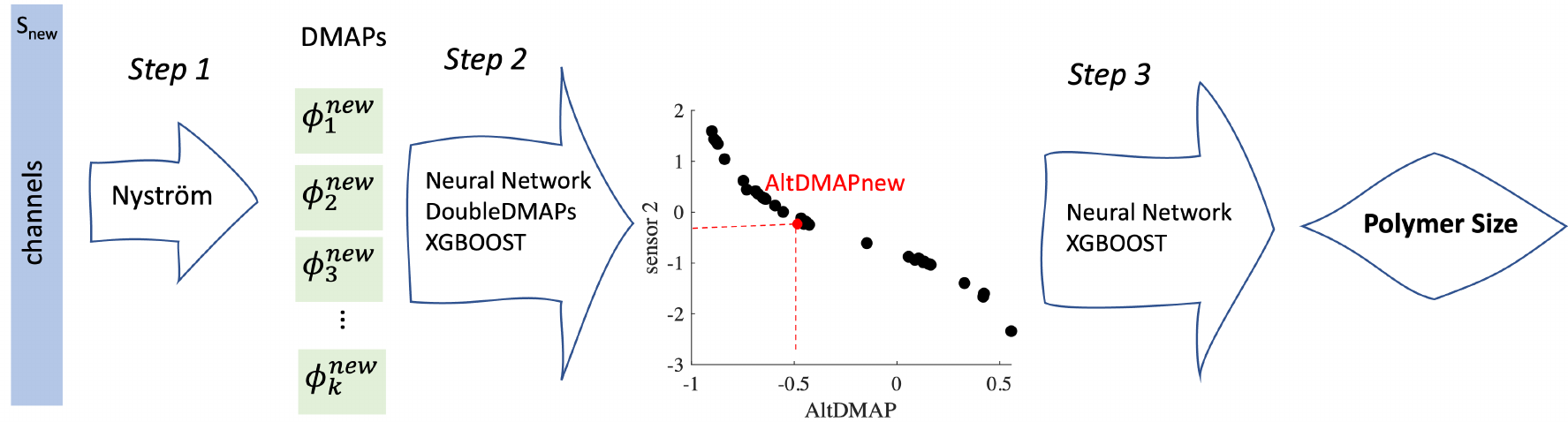}
		\caption{Schematic of the \textit{online} application of size prediction: Given a measured (new) spectrum, the reduced description via DMAPs is determined (Step 1).
			Then, the common AltDMAP variable is inferred with NNs, DoubleDMAPs, or XGBOOST networks (Step 2). 
			Finally, the polymer size is predicted from the common variable, with NNs or XGBOOST (Step 3).}
		\label{fig:online}
	\end{figure}

	The proposed machine-learning workflow consists of two \textit{offline} learning steps, summarized in Fig.~\ref{fig:offline}, that take place once and create the tools for prediction, and three \textit{online} application steps, shown schematically in Fig.~\ref{fig:online}, that are implemented in line with the actual process for fast predictions.
	
	
	
	The two offline steps are:
	\begin{itemize}
		\item Offline Step 1: Dimensionality reduction of the original collection of spectra, consisting of $\sim$~11,000 wavenumbers via DMAPs.
		The goal is to reduce the number of variables to ideally $<$ 10 and thus to re-state the high-dimensional data set in a low-dimensional coordinate system, parameterized by a small number of selected eigenvectors of the kernel matrix defined in equation~\eqref{eq:kernel}.
		The eigenvectors can be selected with the help of the local linear regression algorithm. 
		In this work the eigenvectors are selected based on how accurately the high-dimensional variables are reconstructed, as quantified by an $L^2$-norm of the difference between predicted and actual spectra.
		\item Offline Step 2: Parameterizing the effect of polymer size on the spectra with the AltDMAPs algorithm.
		In this implementation of AltDMAPs, the spectra and the polymer size measurements are considered as the two independent sensor measurements ($s^{(1)}$ and $s^{(2)}$), respectively. 
		The outcome of this algorithm represents the common variable(s) between the two modalities, i.e., the polymer size.
	\end{itemize} 
	
	Having established a reduced representation of the spectra and the common variable between the spectra and the size measurements, we proceed with the \textit{online} prediction steps for a new spectrum.
	A schematic representation of the \textit{online} workflow for size prediction from spectra is presented in Fig.~\ref{fig:online}.
	To this end, three steps are proposed, designed to yield very fast (practically instantaneous) predictions, without further tuning of the model parameters, in an automated manner:
	
	\begin{itemize}
		\item Online Step 1: Compute the DMAP coordinates of a new spectrum using the Nyström extension. 
		\item Online Step 2: Predict the common AltDMAPs variable(s) from the DMAP coordinates of the new spectrum:
		\begin{equation}
			\text{AltDMAP}_i=f_{AltD}(\text{DMAPs}_i)
			\label{eq:DMAPS2Alt}
		\end{equation}
		\noindent
		$i$ here is the i-th new spectrum. 
		The function $f_{AltD}$ can be approximated by various methods, e.g., a fully connected NN, DoubleDMAPS, or XGBOOST.
		\item Online Step 3: The polymer size, $D_\mathrm{H}^\mathrm{predicted}$ is predicted as a function of the AltDMAP coefficients as:
		\begin{equation}
			D_\mathrm{H,i}^\mathrm{predicted}=f_{size}(\text{AltDMAP}_i)
			\label{eq:Alt2size}
		\end{equation}
		\noindent 
		the function $f_{size}$ can be approximated by a feed-forward NN or XGBOOST trained with AltDMAPs as input and polymer size as output.
	\end{itemize}

	\subsection*{Conformal Autoencoders: Y-shaped Architectures}
	\label{sec:yshaped}
\begin{figure}[H]
		\centering
		\begin{subfigure}[b]{\textwidth}
			\centering
			\includegraphics[width=0.7\textwidth]{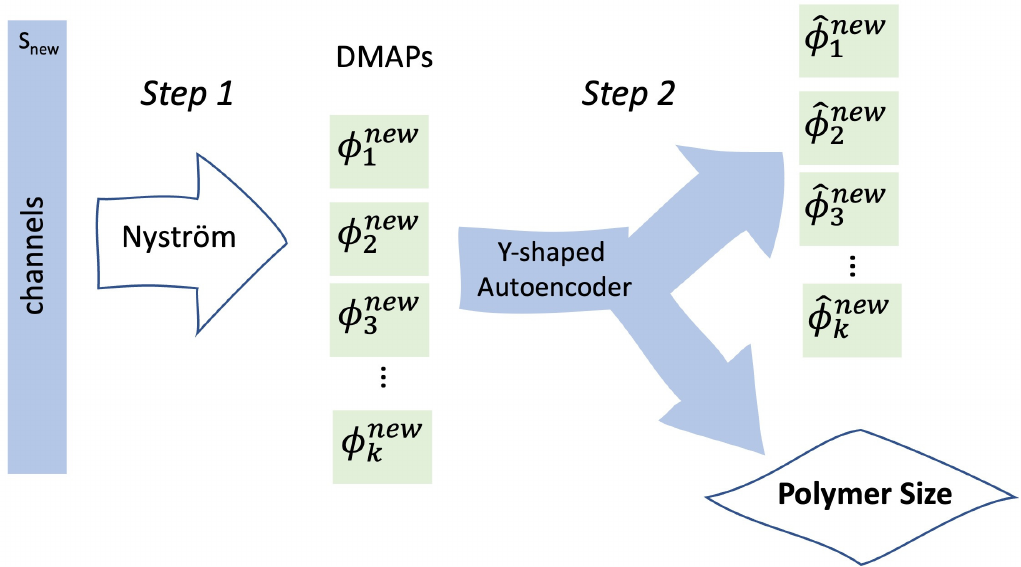}
			\caption{}        
			\label{fig:Yshaped_a}
		\end{subfigure}
		\vfill
		\begin{subfigure}[b]{\textwidth}
			\centering
			\includegraphics[width=0.7\textwidth]{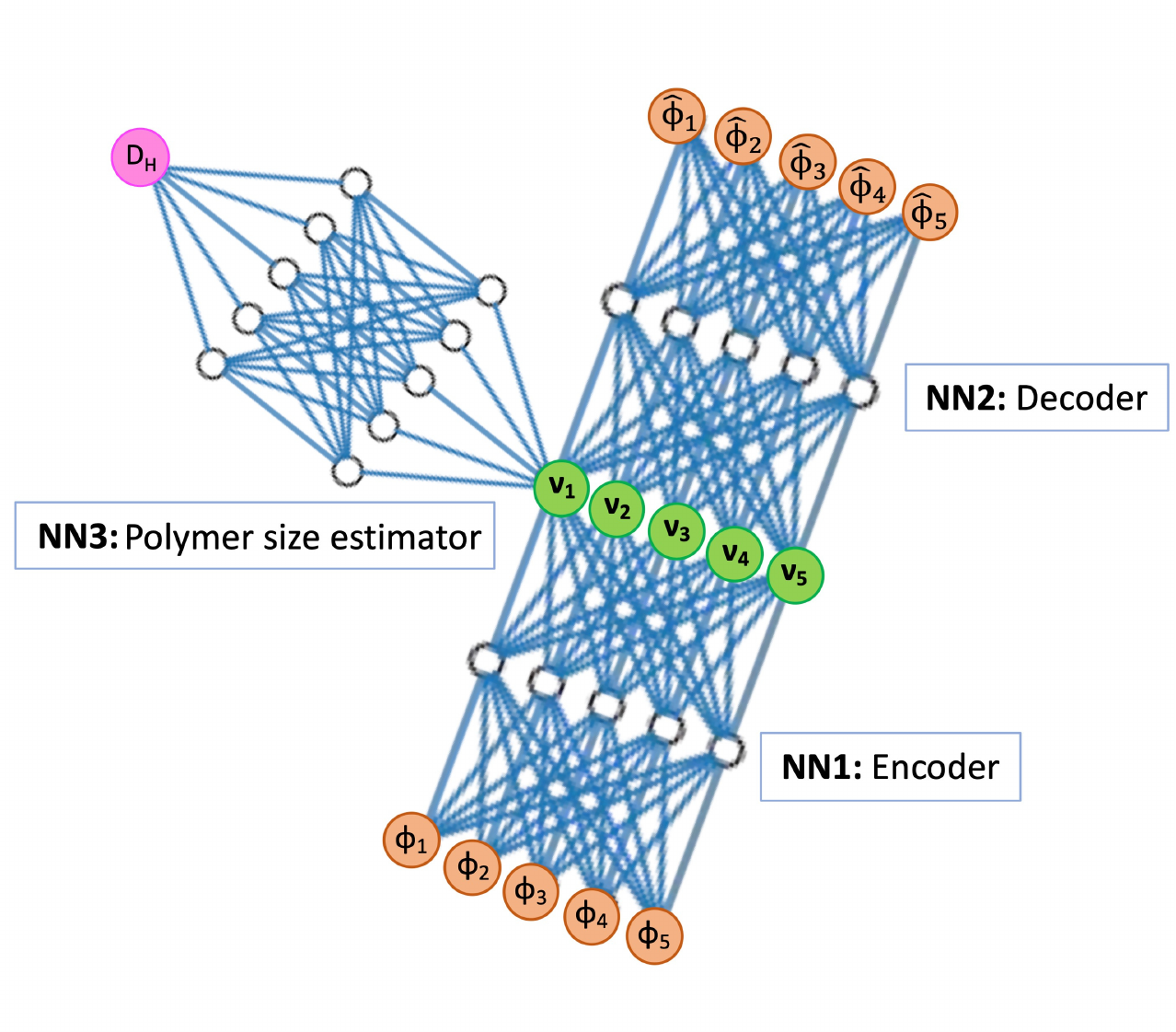}
			\caption{}
			\label{fig:Yshaped_b}
		\end{subfigure}
		\caption{Schematic of the Y-shaped conformal autoencoder architecture: (a) Representation of the workflow; (b) Y-shaped autoencoder composition: NN1 is the encoder that maps the DMAP coordinates to the latent variables of the autoencoder; NN2 is the inverse transformation, from the latent space back to the DMAP coordinates; NN3 maps one of the latent variables to the output of interest: polymer size.}
		\label{fig:Yshape}
	\end{figure}

	Another alternative workflow to predict polymer sizes from Raman spectra involves an ensemble of concurrently trained NNs called Y-shaped conformal autoencoders.
	We use these Y-shaped conformal autoencoders to predict polymer sizes based on DMAP coordinates.
	The schematic representation of the workflow, including the Y-shaped autoencoder, is presented in Fig.~\ref{fig:Yshaped_a}.
	The Y-shaped autoencoder scheme, initially proposed in~\citep{yshaped2022}, is summarized here as it was adjusted for the current application.
	More details on the implementation and training of this machine learning technology are presented in \citep{yshaped2022}.
	At the core of the scheme lies a regular autoencoder, i.e., a NN where the inputs and outputs are the same, with the addition of an extra ``sideways'' NN component, as explained below.
	Overall, the Y-shaped scheme comprises three connected subnetworks (illustrated in Fig.~\ref{fig:Yshaped_b}): 
	\begin{itemize}
		\item Encoder, NN1, which maps the DMAP coordinates, $\phi_i$, to the autoencoder latent variables, $\nu_i$: \\$(\phi_1,\phi_2,\phi_3,\phi_4,\phi_5)\mapsto (\nu_1,\nu_2,\nu_3,\nu_4,\nu_5)$
		\item Decoder, NN2, which can be thought of as the inverse transformation from the latent space of the autoencoder ($\nu_i$) back to the DMAP coordinates $\phi_i$:\\
		$(\nu_1,\nu_2,\nu_3,\nu_4,\nu_5) \mapsto (\hat{\phi}_1,\hat{\phi}_2,\hat{\phi}_3,\hat{\phi}_4,\hat{\phi}_5)$
		\item Polymer size estimator, NN3, which maps the right number of autoencoder latent variables, here one of them, $\nu_1$, to the observed output data, here the polymer size:\\
		$(\nu_1)\mapsto D_\mathrm{H}$
	\end{itemize}
	
	
	The key feature is the loss function, consisting of several parts.
	Successful reconstruction of the input original parameters (the autoencoder part) constitutes the first part. 
	Noteably, the reconstructed inputs, i.e., the output of NN2, are not required further down in our analysis.
	Nevertheless, the accurate reconstruction of the inputs is important for the prediction step, because it ensures that the bottleneck variables, of which one is used for prediction, form a low-dimensional embedding of the data.
	This implies that the bottleneck variables are indeed a parsimonious representation of the original high-dimensional spectra. 
	In theory, this autoencoder part of the NN architecture could also be used for dimensionality reduction of the original high-dimensional data by appropriately pre-selecting the size of the bottleneck layer. 
	However, in the presented case study, the high dimension of the original data set (each spectrum contains $\sim$~11,000 wavenumbers) and the relatively limited number of data-points conclude the preferred strategy to first reduce the dimensionality with DMAPs and subsequently implement the neural network with less variables.
	
	In the following step, comes the ability of NN3, whose input is the single latent variable, here $\nu_1$ is chosen, to reproduce the observed output, i.e., the polymer size; this defines the polymer size as a function of single input, $\nu_1$.
	To concurrently train the different NNs, an additional component to the loss function becomes necessary, which results from further imposing an orthogonality constraint on the conformal autoencoder’s latent coordinates:
	\begin{equation*}
		\langle d\nu_i, d\nu_j \rangle =0,  \forall i\neq j
	\end{equation*}
	
	\noindent{where} $d\nu_i$ indicates the vector of partial derivatives of the latent coordinate $\nu_i$ with respect to of the input parameters $\phi_i$ and $\langle \cdot,\cdot \rangle$ indicates the inner product.
	This constraint is imposed using the automatic differentiation capabilities of the relevant code libraries and aims to disentangle the combination of features that matters to the output from those combinations of features that do not affect it. 

	\section*{Results}
	
	The results comprise the analysis of the size prediction from the benchmark methods, and the developed workflow, including predictions directly from DMAP coordinates via a NN or XGBOOST algorithm, implementation of AltDMAPs and of the Y-shaped autoencoder.
	The Raman spectra and size predictions from DLS measurements of microgel samples are available for transparency~\citep{dataKaven.2023}.
	The codes implementing the different workflow steps can be found in the GitLab repository~\citep{Gitlab_DMAPs.2023}.
	
	\subsection*{Partial Least Squares Regression of Spectral Intensities} \label{sec:results_direct_PLS}
	
	The exhaustive evaluation of various combinations of pretreatment methods for Raman spectra as the basis for PLS regression is summarized in Tab.~\ref{tab:results_PLS}.
	Overall, the direct application of PLS regression to the spectral intensities results in poor prediction performances for any pretreatment method.
	The poor performance is indicated by the R\textsuperscript{2} values significantly lower than 1 for the training and testing.
	Even R\textsuperscript{2} values below zero are encountered.
	Note that R\textsuperscript{2} values below zero imply that the prediction would be more accurate using the mean value than the value predicted by the regression model.
	Comparing the performance of spectra in the FP and global spectral region yields that the prediction performance is independent of the spectral region used.
	For pretreatment involving MinMax normalization in combination with either linear fit or rubber band subtraction or solely linear fit or rubber band subtraction, the number of latent variables needed for the FP region is consistently higher than for the global region, although the global region comprises more prediction variables.
	In contrast, for spectra without pretreatment or with pretreatment involving SNV normalization, the number of latent variables needed for the FP region is lower than for the global region throughout.

\begin{figure}[htp]
		\centering
		\begin{subfigure}{0.45\textwidth}
			\includegraphics[width=\textwidth]{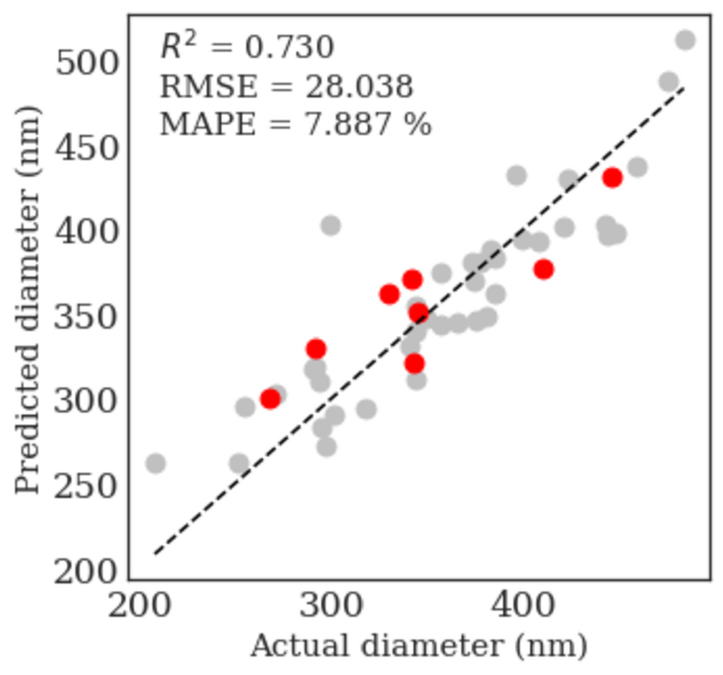}
			\centering
			\caption{Fingerprint region}
			\label{fig:parity_PLS_fp}
		\end{subfigure}
		\begin{subfigure}{0.45\textwidth}
			\includegraphics[width=\textwidth]{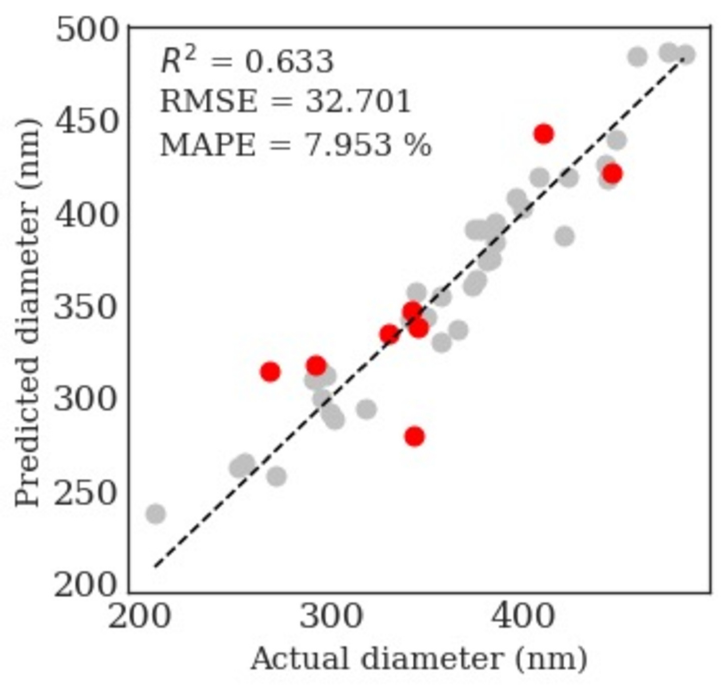}
			\centering
			\caption{Global region}
			\label{fig:parity_PLS_global}
		\end{subfigure}
		\caption{Parity plots of microgel size predictions via PLS regression of raw spectra in (a) the fingerprint region and (b) the global region. 
			Gray circles represent the training data, and red circles indicate the test data.}
		\label{fig:parity_PLS}
	\end{figure}

	
	In summary, the predictions using raw spectra with no pretreatment in the global region perform best in training and testing considering all performance metrices.
	Also, the prediction based on raw spectra in the FP region with no pretreatment shows a relatively good performance indicated by the second-best accumulated R\textsuperscript{2} value (training and testing) of \SI{1.520}{}.
	Thus, the parity plots for these most promising configurations are shown in Fig.~\ref{fig:parity_PLS} in comparison.
	Here, the gray circles represent the training data, and the red circles represent the test data.
	Over-fitting is precluded sufficiently, as the discrepancy between actual and predicted size is in the same range for the training and test data set.
	Furthermore, comparing the PLS results based on the raw spectra in the FP region (Fig.~\ref{fig:parity_PLS_fp}) and the global region (Fig.~\ref{fig:parity_PLS_global}) shows no significant improvement in the PLS performance caused by the spectral range. 
	However, the limited size of the data set is probably a restrictive factor and with an extended data set the performance of the state-of-the-art methods might improve.
	Therefore, the findings regarding the application of the state-of-the-art methods to the employed data set can not be generalized.

	\subsection*{Regression of Hard Model Parameters} \label{sec:results_IHM_PLS}
	
	In Tab.~\ref{tab:results_IHM_PLS}, we show the overview of different pretreatment methods in the indirect hard modeling step.
	The fitted parameter values from the IHM are subsequently used in the PLS regression.
	We considered a high and medium fitting mode corresponding to more or less degrees of freedom for fitting the IHM evaluation model to the experimental spectra.
	Similarly to the direct regression on spectral intensities, we find that the overall performance of the predictions is unsatisfying.
	Again, we determine R\textsuperscript{2} values significantly lower than 1 for the training and even below zero for testing in some cases.
	Overall, the medium fitting mode shows a poorer prediction performance than the high fitting mode.
	With respect to the latent variables, for pretreatment involving rubber band more latent variables are needed in high fitting mode than in medium fitting mode.
	In contrast to the expected outcome that the high fitting mode necessitates more latent variables, as we need to reduce a larger space of prediction variables, the remaining pretreatment methods not involving rubber band subtraction result in less latent variables for high than for medium fitting mode.
	Furthermore, no clear trend exists that one aspect of the pretreatment method performs better than the other.
	
\begin{figure}[htp]
		\centering
		\begin{subfigure}{0.45\textwidth}
			\includegraphics[width=\textwidth]{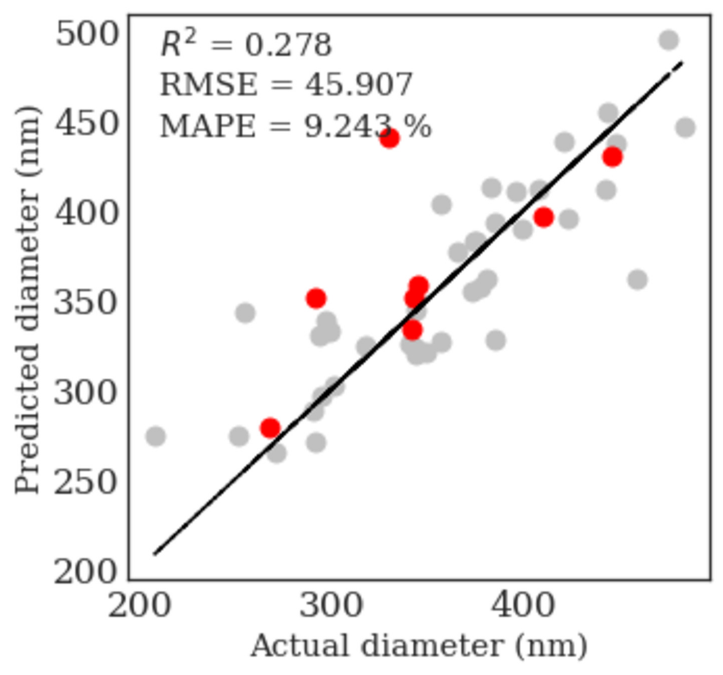}
			\centering
			\caption{MinMax, linear fit}
			\label{fig:parity_IHM_PLS_Minmax_linfit_high}
		\end{subfigure}
		\begin{subfigure}{0.45\textwidth}
			\includegraphics[width=\textwidth]{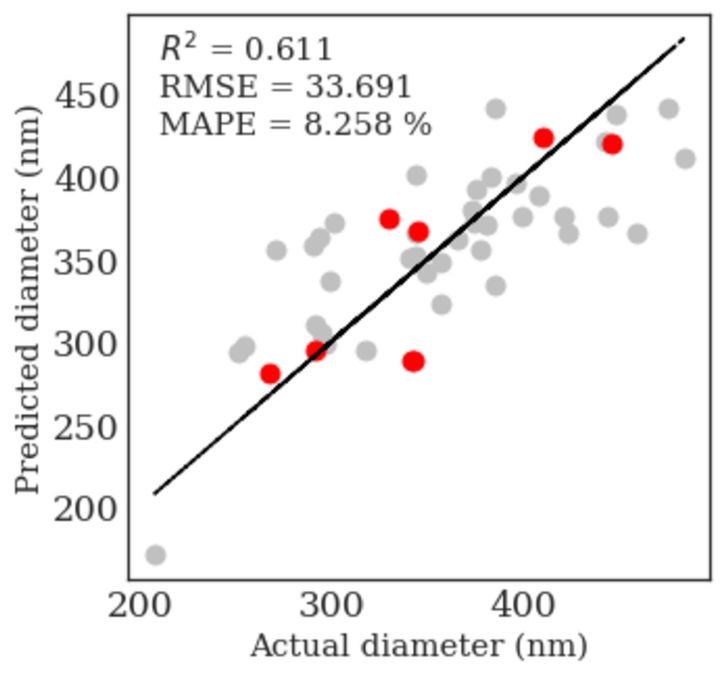}
			\centering
			\caption{SNV, rubber band}
			\label{fig:parity_IHM_PLS_SNV_rubber_high}
		\end{subfigure}
		\caption{Parity plots of microgel size predictions via PLS regression of IHM parameters from spectra fitted via high interaction.
			Gray circles represent the training data, and red circles indicate the test data.}
		\label{fig:parity_IHM_PLS}
	\end{figure}

	In Fig.~\ref{fig:parity_IHM_PLS}, we present the results of the PLS regression based on IHM parameters from Raman spectra pretreated via SNV and rubber band baseline subtraction and high interaction during the fitting, as this configuration yields the relatively best performance according to Tab.~\ref{tab:results_IHM_PLS}.
	Also, we show the second best performing configuration, namely regression based on spectra pretreated with MinMax normalization and a linear fit subtraction and fitted with high interaction.
	Similarly to the results from the direct application of PLS, the distribution of gray circles~(training data) and the red circles~test data) here shows that over-fitting is suppressed.
	In addition, the comparison of the PLS results based on the IHM parameter values from SNV plus linear fit pretreated spectra (Fig.~\ref{fig:parity_IHM_PLS_SNV_rubber_high}) and spectra pretreated with MinMax normaliztion and a linear fit subtraction (Fig.~\ref{fig:parity_IHM_PLS_Minmax_linfit_high}) shows no notable improvement for the spectra pretreated via SNV in the PLS performance. 
	In conclusion, the resulting prediction performance does not indicate reliable prediction accuracy.
	The reduction of the spectral range to IHM parameters constitutes a physically meaningful approach, but is shown to be insufficient for a limited size of data sets.
	For IHM in combination with PSL, the performance might improve with an extended data set and the findings can hence not be generalized but apply solely to the employed data sets.


	\subsection*{Prediction Directly from Diffusion Maps} \label{sec:results_DMAPs}
\begin{figure}[H]
		\centering
		\includegraphics[width=0.7\textwidth]{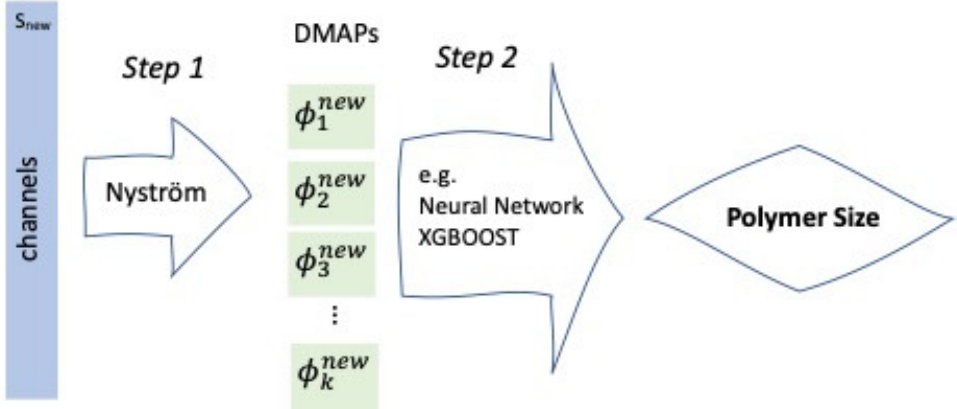}
		\caption{Schematic representation of the prediction strategy directly from the DMAP coordinates that parsimoniously parameterize the spectra.}
		\label{fig:onlineDirecltyfromDMAPs}
	\end{figure}

	Firstly, the DMAPs algorithm is implemented, and six DMAP coordinates, $\phi_1$, $\phi_2$, $\phi_3$, $\phi_4$, $\phi_5$, and $\phi_6$ are selected to parsimoniously represent the spectra that live in a high-dimensional ambient space (cf. Fig.~\ref{fig:onlineDirecltyfromDMAPs}, Step 1). 
	These are selected based on how accurately the original data can be reconstructed from those latent variables. 
	Here the $L^2$ norm of the difference between the predicted and actual spectra, for the test set, is $3.34 \cdot 10^{-6}$.
	Subsequently, the polymer size is directly predicted from DMAPs coordinates (cf. Fig.~\ref{fig:onlineDirecltyfromDMAPs}, Step 2). 
	
\begin{figure}[htp]
		\centering
		\begin{subfigure}{0.45\textwidth}
			\includegraphics[width=\textwidth]{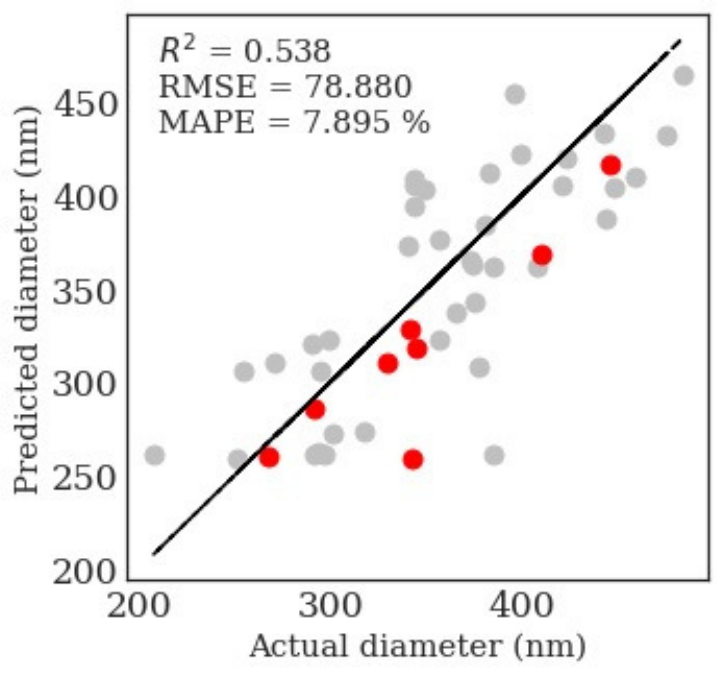}
			\centering
			\caption{}
			\label{fig:simpleNN}
		\end{subfigure}
		\begin{subfigure}{0.45\textwidth}
			\includegraphics[width=\textwidth]{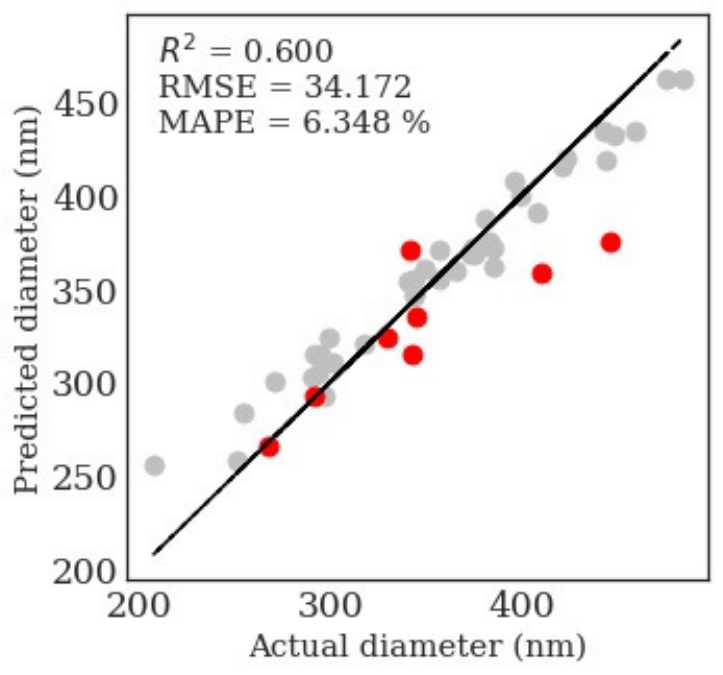}
			\centering
			\caption{}
			\label{fig:simpleXGBOOST}
		\end{subfigure}
		\caption{Parity plots of microgel size predictions from DMAPs (a) by a neural network and (b) using XGBOOST. 
			Red points correspond to the test set, and gray points correspond to the training set values. 
			The reported accuracy metrics (R\textsuperscript{2}, RMSE, and MAPE) correspond to the test data.}
		\label{fig:simple}
	\end{figure}

	To this end, a feed-forward NN is trained, having as inputs the DMAP coordinates that correspond to the training data and, as output, the polymer size. The hyper-parameters of the network here and in subsequent sections are fine-tuned using the $KerasTuner$ $RandomSearch$  hyper-parameter optimization framework.
	The results of direct predictions from DMAP coordinates are presented in Fig.~\ref{fig:simple}. 
	It is possible to achieve regression, with MAPE$=7.895\%$ and R\textsuperscript{2}$=0.538$, for the test set (cf. Fig.~\ref{fig:simpleNN}), and it is more challenging to identify the hyper-parameters that prevent over-fitting. 
	Alternatively, the XGBOOST algorithm is used with randomized search on hyper-parameters of the XGBOOST estimator, using $RandomizedSearchCV$ from the scikit-learn library in Python. 
	The parameters of the estimator are optimized, here and in the implementations presented in subsequent sections by cross-validated search over parameter settings. 
	For efficiency, not all parameter values are tested, but rather a fixed number of parameter settings is sampled from the specified distributions. 
	The performance of the method has similar accuracy to the neural networks (cf. Fig.~\ref{fig:simpleXGBOOST}, MAPE$=6.348\%$ and R\textsuperscript{2}$=0.600$).

	\subsection*{Prediction using Alternating Diffusion Maps} \label{sec:results_AltDMAPs}
	
\begin{figure}[H]
		\centering
		\begin{subfigure}[b]{0.45\textwidth}
			\centering
			\includegraphics[width=\textwidth]{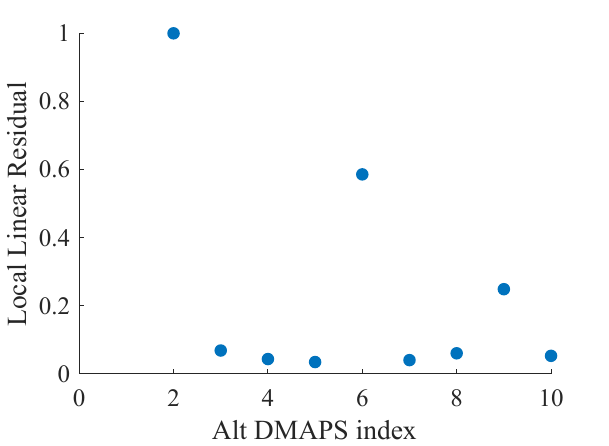}
			\caption{}
			\label{fig:llr}        
		\end{subfigure}
		\begin{subfigure}[b]{0.47\textwidth}
			\centering
			\includegraphics[width=\textwidth]{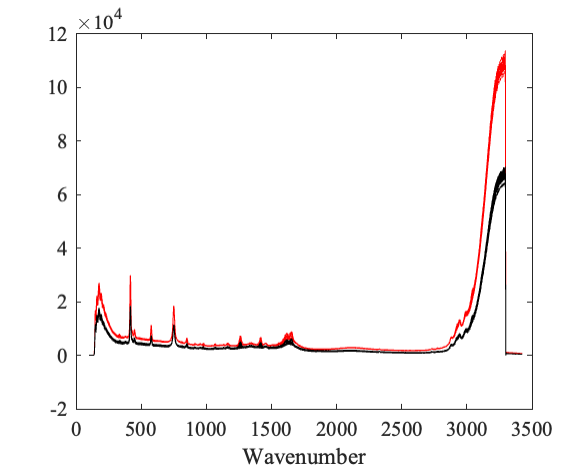}
			\caption{}
			\label{fig:originalspectra}        
		\end{subfigure}
		\begin{subfigure}[b]{0.47\textwidth}
			\centering
			\includegraphics[width=\textwidth]{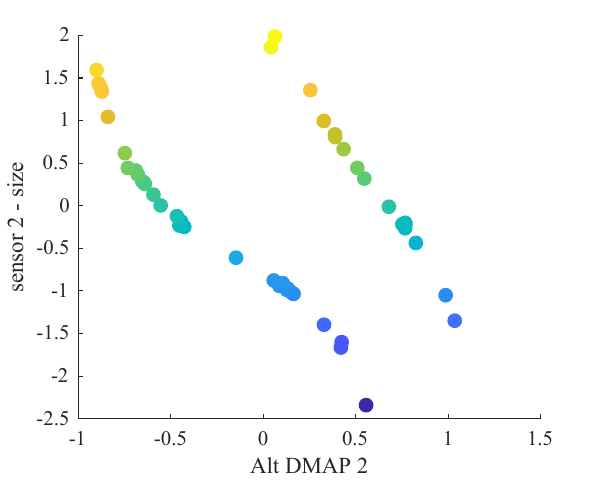}
			\caption{}
			\label{fig:altvss2}        
		\end{subfigure}
		\caption{(a) Local linear regression indicates two significant common variables exist; (b) Plot of the raw spectra included in the data set; two clusters appear with respect to the spectral intensities, colored in red and black; (c) The (normalized) polymer size is plotted over the one common variable (AltDMAP) for both clusters in (b) illustrating the dependence of size on the latent variables (one-to-one within each individual cluster). 
			To be able to predict the size from any spectrum, irrespective of the the cluster it belongs to, more than one AltDMAP is required.
		}
		\label{fig:AltDMAP}
	\end{figure}

	The AltDMAPs algorithm is implemented to find common variables, where the collection of spectra is considered as the $s^1$ (Sensor 1) measurements and the corresponding polymer sizes as $s^2$ (Sensor 2).
	Two significant common variables at AltDMAPs index 2 and 6 are found by implementing the local linear regression algorithm, indicated by a high value of the local linear residual in Fig.~\ref{fig:llr}.
	

	The common variables parameterize the spectra variability attributed to the polymer size. 
	It is worth noting that two clusters appear in the original data set of raw spectra concerning the spectral intensities (cf. Fig.~\ref{fig:originalspectra}). 
	The common variable AltDMAP 2 is one-to-one with the size in each cluster separately, as visually demonstrated, in Fig.~\ref{fig:altvss2}. 
	Overall, for both clusters, the size is not a function of a single AltDMAP, and the AltDMAP variable with index $6$ appears to be correlated to the polymer size, to some extent, since the local linear residual value (cf. Fig.~\ref{fig:llr}) is non-negligible. 
	
\begin{figure}[H]
		\centering
		\begin{subfigure}[b]{0.45\textwidth}
			\centering
			\includegraphics[width=\textwidth]{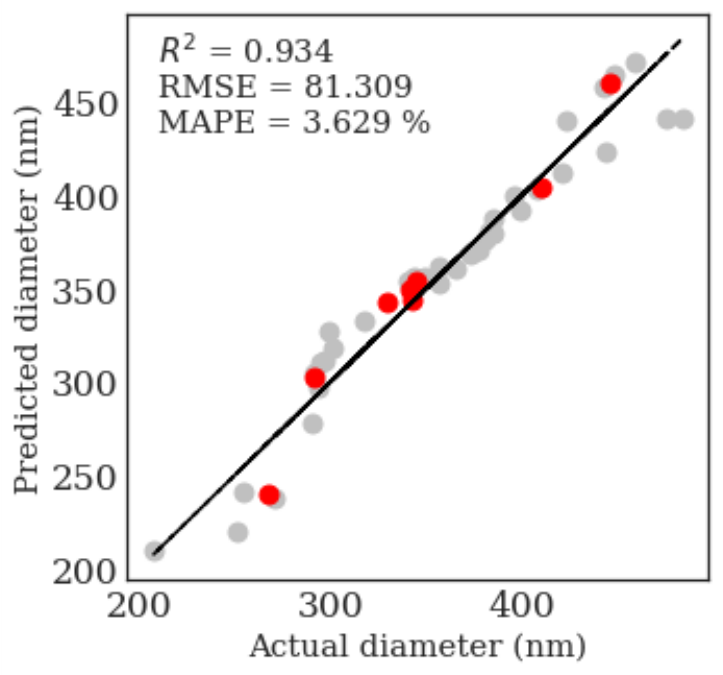}
			\caption{}
			\label{fig:DiamFromNN}        
		\end{subfigure}
		\begin{subfigure}[b]{0.45\textwidth}
			\centering
			\includegraphics[width=\textwidth]{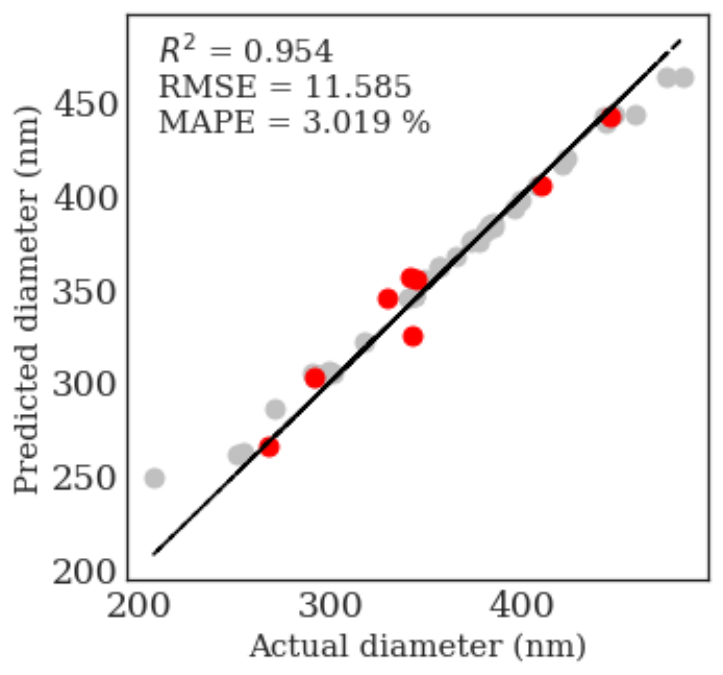}
			\caption{}
			\label{fig:DiamFromXGBOOST}        
		\end{subfigure}
		\caption{Parity plots of microgel size predictions from AltDMAPs, with (a) NN, and (b) XGBOOST algorithm.
			The gray points correspond to the training set data, and the red points correspond to the test data. 
			The reported accuracy metrics correspond to the test data.}
		\label{fig:DiamFromAlt}
	\end{figure}

	Identifying a reduced description of the spectra and common variables between the spectra and the polymer size enables predicting the polymer size corresponding to a new spectrum by following the three \textit{online} steps in the workflow (cf. Fig.~\ref{fig:online}).
	As part of Online Step 1, the DMAP coordinates of a new set of spectra are determined by implementing the Nyström extension.
	The DMAP coordinates are accurately predicted by the Nyström extension, achieving a mean squared error, MSE$=1.58\times10^{-7}$.
	Subsequently, in Online Step 2, the common variables, AltDMAPs, are computed as a function of the respective DMAP coordinates of the test set computed in the previous step.
	Two alternatives are implemented: Double Diffusion Maps, with a prediction error of MSE$=0.0248$, and the Extreme Gradient Boosting (XGBOOST) algorithm, which is slightly more accurate here, achieving MSE$=0.0162$. 
	In practice, including more AltDMAP coordinates for a new spectrum is necessary, here we include the first six AltDMAPs. 
	The final step of the workflow involves predicting the polymer size given the common variables, AltDMAPs.
	The prediction results using a feed-forward NN and XGBOOST are shown in Fig.~\ref{fig:DiamFromAlt}.
	Here, the input is the AltDMAP coordinates, and the output is the polymer size.
	Both methods yield similar results regarding MAPE, RMSE, and R\textsuperscript{2} for the test set.
	
\begin{figure}[H]
		\centering
		\begin{subfigure}[b]{0.45\textwidth}
			\centering
			\includegraphics[width=\textwidth]{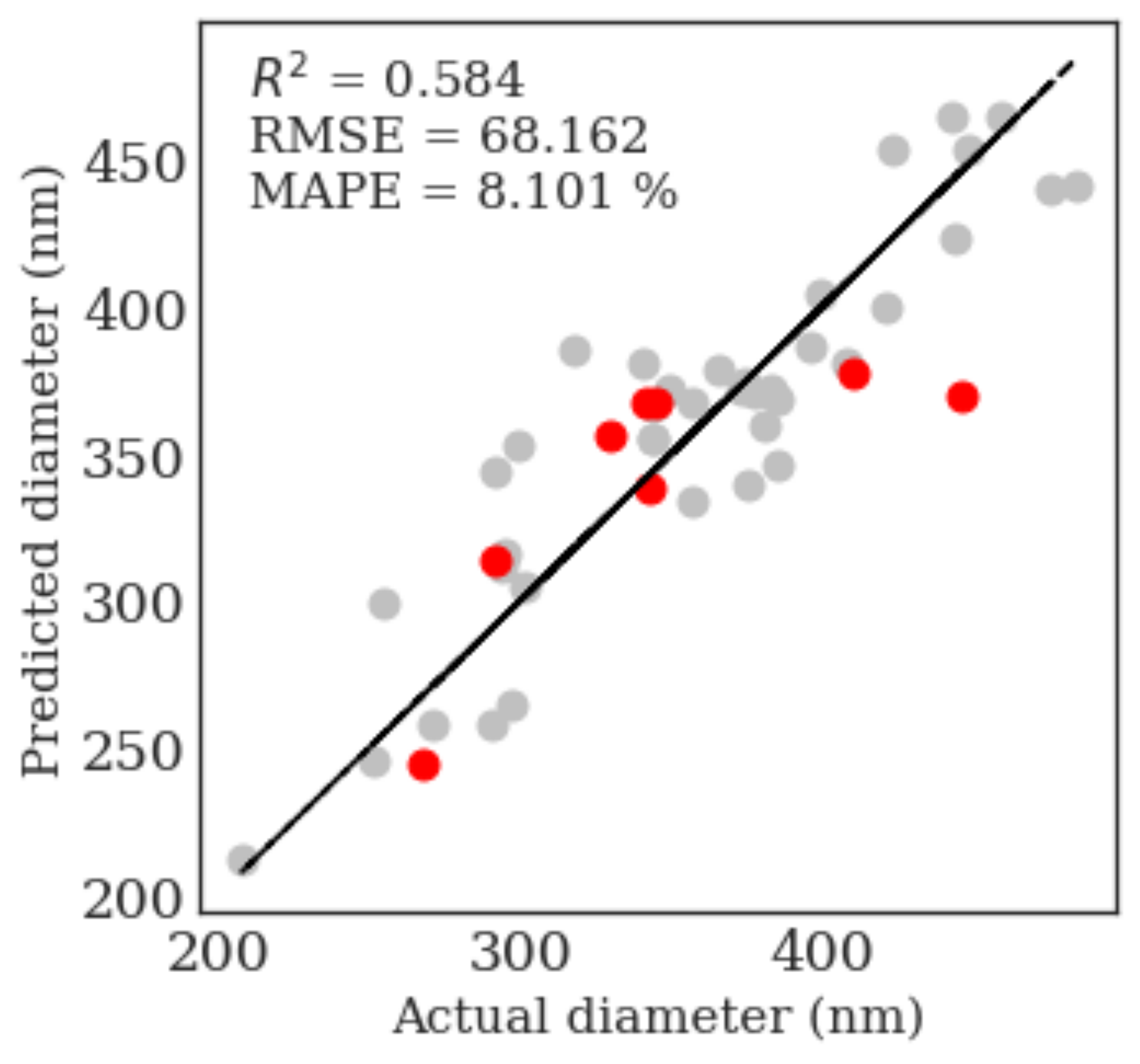}
			\caption{}
			\label{fig:SizeNNAltDouble}        
		\end{subfigure}
		\begin{subfigure}[b]{0.45\textwidth}
			\centering
			\includegraphics[width=\textwidth]{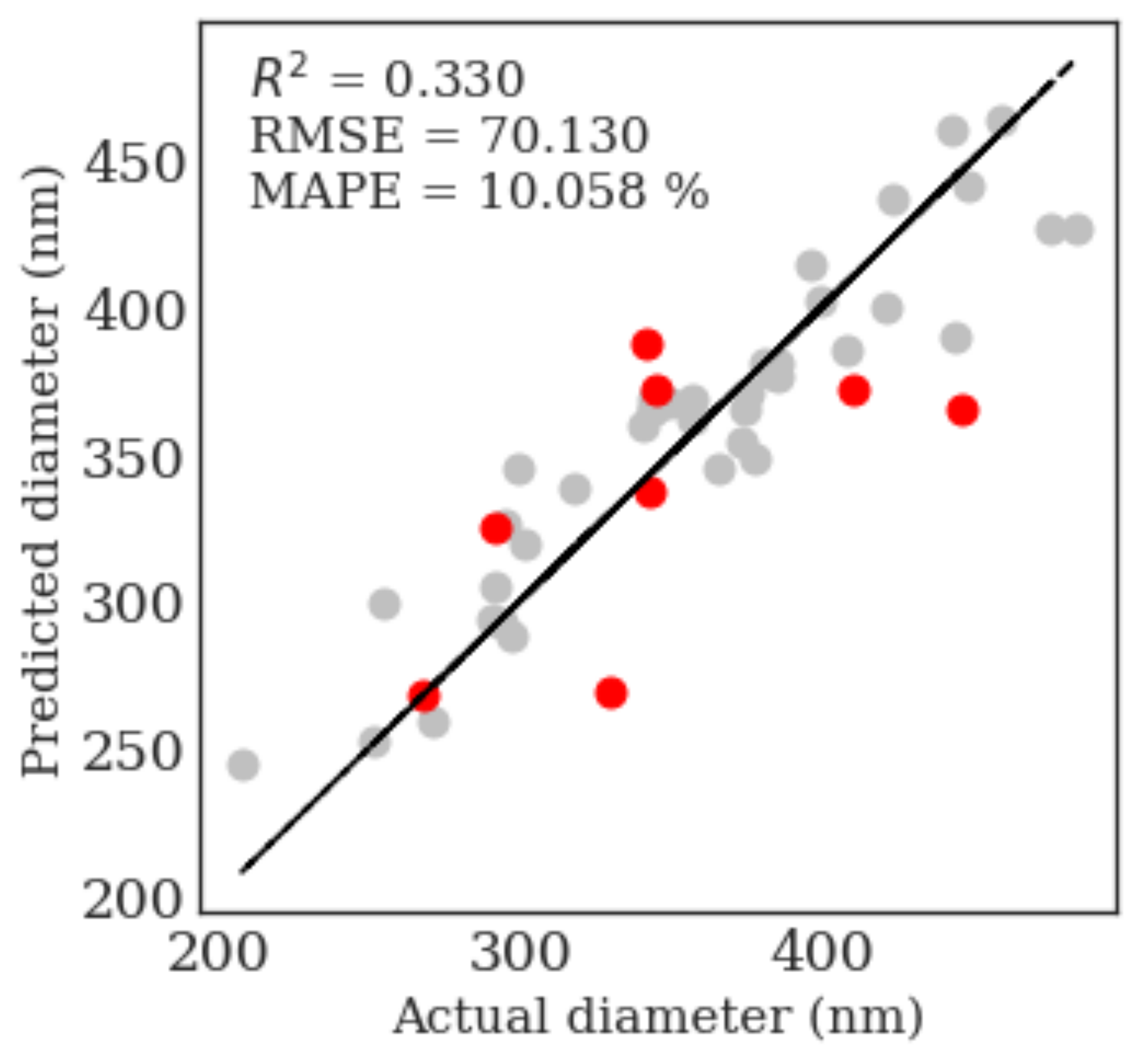}
			\caption{}
			\label{fig:SizeNNAltXGBOOST}        
		\end{subfigure}
		\begin{subfigure}[b]{0.45\textwidth}
			\centering
			\includegraphics[width=\textwidth]{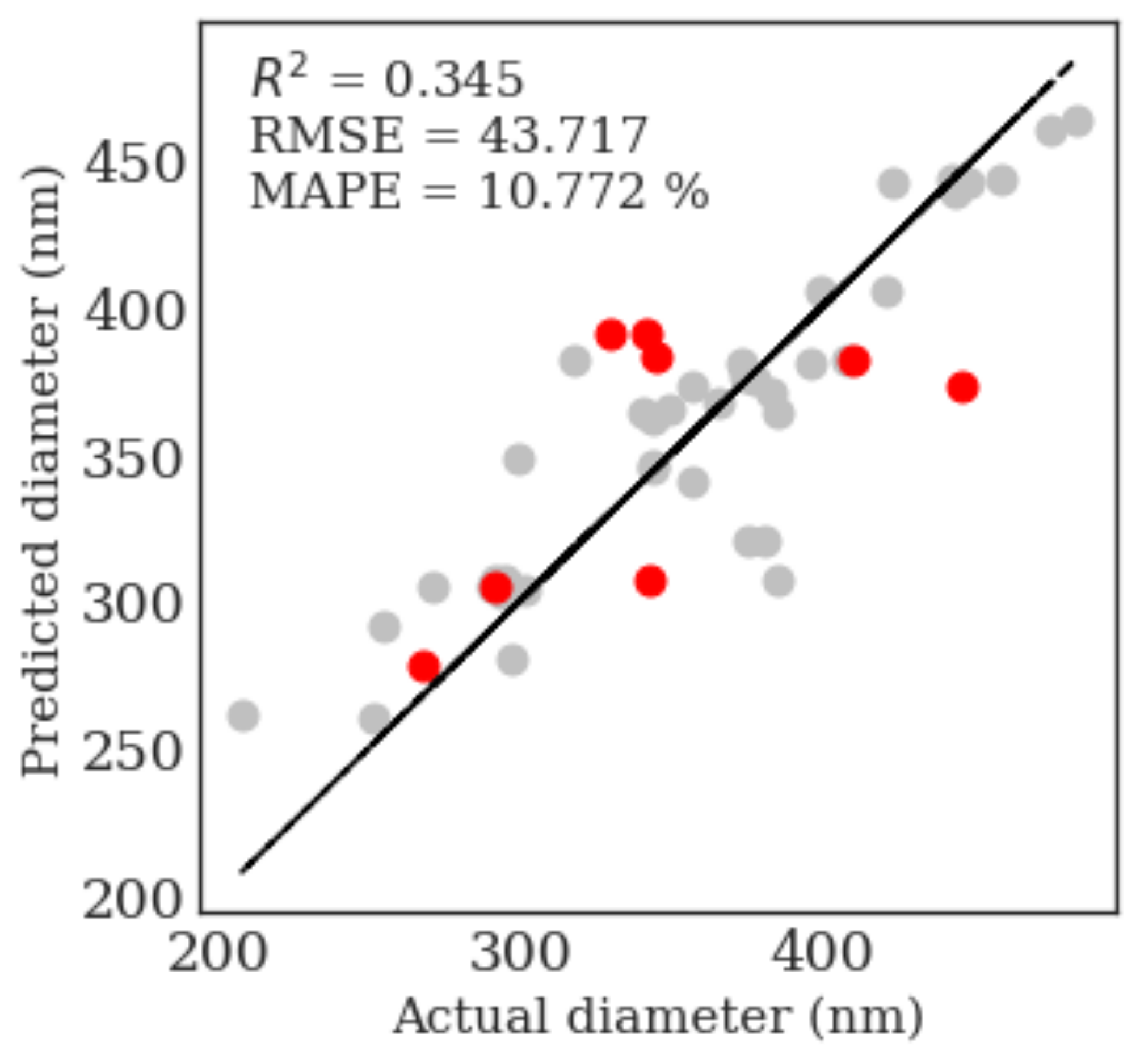}
			\caption{}
			\label{fig:SizeXGBOOSTAltDouble}        
		\end{subfigure}
		\begin{subfigure}[b]{0.45\textwidth}
			\centering
			\includegraphics[width=\textwidth]{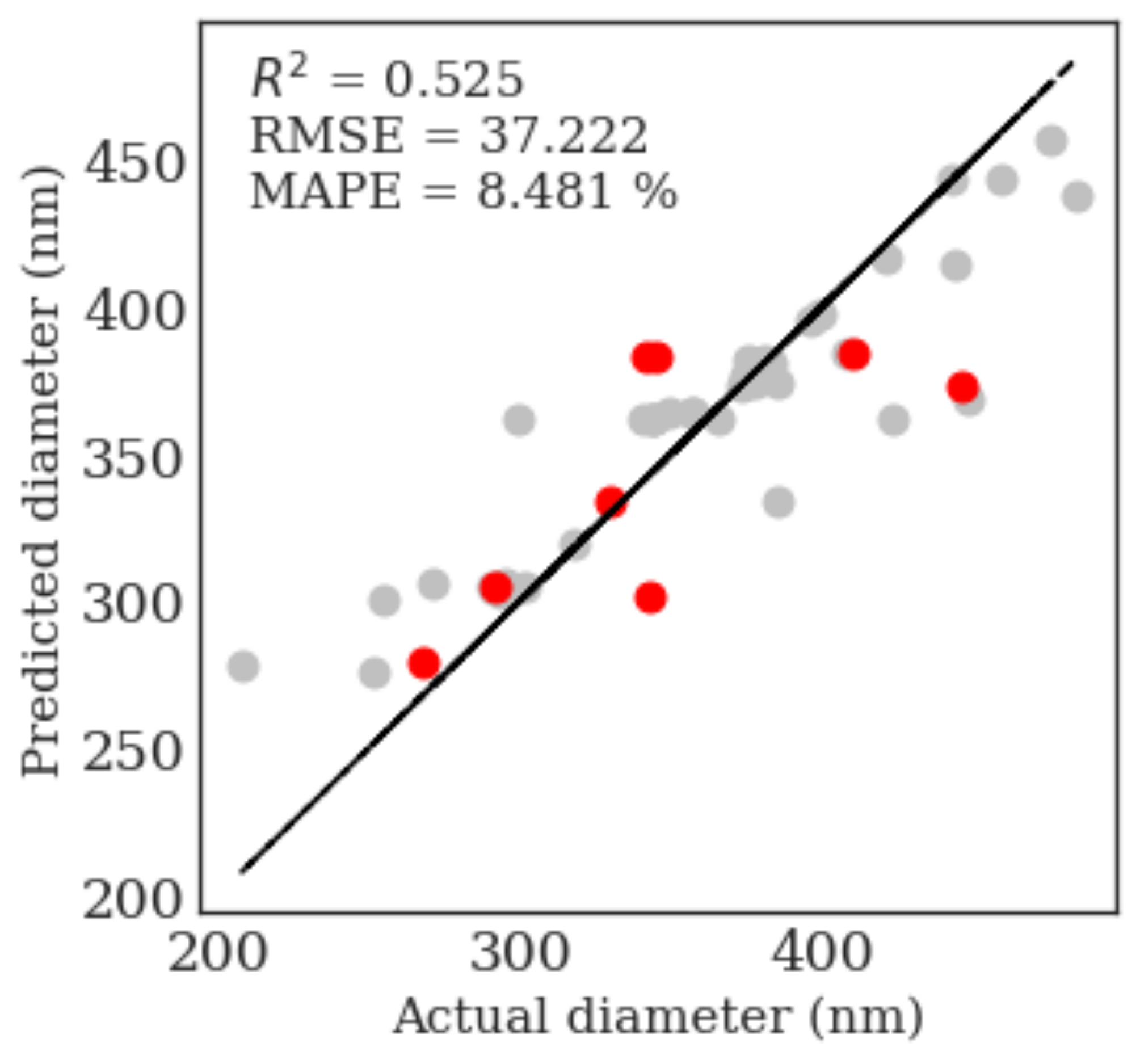}
			\caption{}
			\label{fig:SizeXGBOOSTAltXGBOOST}        
		\end{subfigure}
		\caption{Parity plots of microgel diameter predictions via (a) a NN from DoubleDMAPs-predicted AltDMAPs, (b) a NN from XGBOOST-predicted AltDMAPs, (c) XGBOOST from Double DMAPs-predicted AltDMAPs, and (d) XGBOOST from XGBOOST-predicted AltDMAPs.
			The gray points correspond to the training set data, and the red points correspond to the test data.
			The reported accuracy metrics (R\textsuperscript{2}, RMSE, and MAPE) correspond to the test data.}
		\label{fig:AltD2Size}
	\end{figure}

	
	The performance of the overall \textit{online} workflow is shown in Fig.~\ref{fig:AltD2Size}.
	Specifically, four combinations of methods are implemented and reported: (i) the size is predicted by a NN from AltDMAPs predicted by Double DMAPs, leads to an R\textsuperscript{2}$=0.584$ and MAPE$=8.101\%$ for the test set; (ii) the size is predicted by a NN from AltDMAPs predicted by XGBOOST, leads to an R\textsuperscript{2}$=0.330$ and MAPE$=10.058\%$ for the test set; (iii) the size is predicted by the XGBOOST algorithm from AltDMAPs predicted by Double DMAPs, leads to an R\textsuperscript{2}$=0.345$ and MAPE$=10.772\%$ for the test set; (iv) the size is predicted by the XGBOOST algorithm from AltDMAPs predicted by XGBOOST, leads to an R\textsuperscript{2}$=0.525$ and MAPE$=8.481\%$ for the test set. 
	As shown, the prediction of the polymer particle diameter from the \textit{predicted} AltDMAPs is less accurate than the prediction of the \textit{actual} AltDMAPs, which is attributed to the prediction error of the AltDMAPs from the DMAPs. 
	Overall, case (i) is the best-performing one for the available data set for R\textsuperscript{2} and MAPE. 
	Considering the $\%$-error in size prediction, the combination of methods of case~(i) is more advantageous regarding the maximum absolute values of the $\%$-error and error distribution.
	
	

	Overall, the predictive accuracy directly from DMAPs is slightly better than when implementing the AltDMAP methodology, although they are characterized by the same order of magnitude. 
	Given the size of the available data set and the difference between the performance metrics (R\textsuperscript{2}, MAPE), it is unclear whether this small difference is meaningful enough to persist given a different or larger data set.
	
	\subsection*{Y-shaped Autoencoder} \label{sec:results_autoencoder}
	
	The conformal autoencoder architecture is trained with the same training set as the previous methods: 
	The DMAP coordinates $(\phi_1,...,\phi_6)$ corresponding to the training set are the input to the encoder network NN1.
	These values are also the target values for the decoder network NN2.
	We set six latent variables in the bottleneck layer, and we require that the polymer size be defined as a function of $\nu_1$, by concurrently training the neural network NN3.
	
	The performance of this method is superior to both the AltDMAPs workflow and the direct prediction from DMAP coordinates with a R\textsuperscript{2} value of $0.951$ and MAPE value of $2.93\%$ for the test set.
	We believe that the reason for the enhanced performance lies in the fact that the Y-shaped architecture not only finds the latent variables of the data set, as do DMAPs, but with the second NN, specific properties are explicitly imposed on one of them: 
	The polymer size must be a function of one latent variable orthogonal to all other latent variables.
	The latter property implies that only this one latent variable correlates to the particle diameter, which is a subtle but essential difference from the AltDMAPs approach (two common variables and even more required for accurate online prediction). 
	This hypothesis merits further investigation, which lies beyond the scope of this work.
	
	\subsection*{Comparison of Presented Methods}
	
	In Tab.~\ref{tab:comparison_summary}, we compare the results from the state-of-the-art to the proposed methods.
	We cluster the methods in three categories: state-of-the-art, alternating diffusion maps, and prediction directly from DMAP coordinates.
	The results using the respective methods are described in detail in previous sections.
	Here, we present the best-forming configuration of pretreatment and spectral range for the direct PLS regression on Raman spectra and PLS regression on IHM parameters.
	
	For Raman spectra studied here, the Y-shaped autoencoder outperforms the other considered methods of this work indicated by the only R\textsuperscript{2} value considerably close to $1$.
	Also, the low RMSE values of the Y-shaped autoencoder, being approximately a third of the values of the compared methods, indicate improved performance.
	Lastly, the MAPE value of the autoencoder-based method ranges at \SI{2.930}{\percent}, which approximates the precision range of around \SI{2}{\percent} expected from DLS~\citep{Malvern.2006}, the established size measurement device.
	Thus, even though the evaluation is based on a data set with limited size, for the first time, a purely data-driven evaluation method based on Raman spectra advances the accuracy capability of the established size determination device.
	The next best-performing methods in the current study include the prediction directly from DMAP coordinates via the XGBOOST algorithm, the PLS regression from IHM parameters, and the direct PLS regression.
	Each method reaches a similar prediction performance of R\textsuperscript{2} values of \SI{0.600}{}, \SI{0.636}{}, and \SI{0.633}{}, respectively.
	Also, their RMSE values are close to each other, ranging at \SI{34.172}{\nano\meter}, \SI{34.840}{\nano\meter}, and \SI{32.700}{\nano\meter}.
	Finally, the MAPE value of the XGBOOST involving method is with \SI{6.348}{\percent} slightly better than the PLS regression based on IHM parameters with \SI{8.573}{\percent} and the direct PLS regression with \SI{7.953}{\percent}.
	
	All methods involving AltDMAPs only achieve an inferior prediction accuracy with R\textsuperscript{2} values between \SIrange{0.330}{0.584}{}.
	Within the AltDMAPs involving methods, there is no clear trend suggesting which combination of AltDMAPs prediction with size prediction enhances the performance.
	
	On that note, it is worth looking further into the collection of methods that rely on manifold learning (DMAPs, AltDMAPs, and Y-shaped autoencoder), specifically on the number of latent variables required (cf. the last column of Tab.~\ref{tab:comparison_summary}) and discuss their unique characteristics. 
	The first observation is that DMAPs are able to meaningfully embed the high dimensional data using six coordinates. 
	Even though a parsimonious embedding of the data is achieved, which is key in making machine learning tasks computationally tractable, it is not a quantitatively accurate predictor. 
	Here, the ability of the AltDMAPs to identify the common variables between the measured observation (the size) and the low dimensional description of the input (Raman spectra) by the DMAP coordinates is beneficial.
	Consequently, we further reduce the variables that are meaningful for the specific task of predicting polymer size from six to two latent variables, on the data. 
	%
	It becomes clear that ``designer" latent spaces, i.e., ones with specific desired characteristics, become useful in disentangling the data features and creating latent variables that specifically map to the observable quantity of choice, here polymer size. 
	This finding motivates the proposed conformal autoencoder architecture.
	The autoencoder network identifies a single latent variable, which not only maps to the size but is independent of other data features by design (imposed orthogonality condition in the loss function of the NN). 
	This trait of the conformal autoencoder architecture enables our accurate prediction of polymer size here.
	

	\section*{Conclusions and Future Work}
	This contribution considers the important open problem of predicting particle sizes online.
	We propose the prediction of monodisperse polymer sizes from Raman spectra via a data-driven approach.
	We apply the approach to our open-access data set of continuous microgel synthesis and demonstrate its capability.
	Further, we compare our approach against two state-of-the-art benchmark methods to underline the excellent prediction performance of our nonlinear approach.
	
	All proposed nonlinear approaches rely on dimensionality reduction via DMAP coordinates.
	We find that the machine learning workflow combining the data reduction capability of the DMAPs algorithm with the recent advances of Y-shaped autoencoder outperforms all alternatively considered workflows significantly.
	The remaining proposed methods with altering configurations of involved algorithms accomplish prediction performances comparable to state-of-the-art linear methods.
	In contrast, the Y-shaped autoencoder approach enables drastically better prediction accuracy, similar to the established size measurement methods such as DLS.
	Therefore, we derive an algorithmic workflow for prediction of particle sizes from in-line measurements that is competitive with off-line analytical methods.
	
	Predicting polymer sizes directly from in-line Raman spectra taken from untreated samples, as labor-intensive DLS processing is circumvented and online reaction monitoring for closed-loop control is enabled.
	Furthermore, with the proposed method the spectra are not manipulated by spectral pretreatment, thus, no expert knowledge is necessary.
	In contrast to established machine learning workflows, the proposed algorithm exhibits high efficiency, as the algorithmic filtering enables a proficient prediction performance based on a data set of limited size.
	This aspect makes the proposed workflow especially relevant, as requiring experimental data is laborious.
	In addition, the workflow typically relies on less than ten coordinates in the reduced component space.
	Here, only the first six DMAP coordinates enable us to use the entire wavelength spectrum without exclusions, which would otherwise require problem-specific intuition.
	
	Future works include the application of the proposed workflow to predict polymer concentrations and sizes simultaneously to highlight the application of our proposed readily available analysis tool.
	The simultaneous prediction allows a more comprehensive characterization via a single in-line process analytical tool.
	Furthermore, subsequent investigations include the open challenge of predicting size distributions from in-line Raman spectroscopy.
	In addition, future considerations involve extending the workflow to other systems beyond the presented application, which can involve crystallization processes, amongst others.

	\section*{Author contributions}
	E.D.K.: developed, implemented, and applied the DM, ADM, and Y-shaped autoencoder frameworks, developed the heuristic pre-processing, wrote initial draft;
	L.F.K.: performed experimental design, preprocessed data, implemented and applied PLS and hybrid PLS+IHM method, wrote initial draft;
	J.M.M.F.: performed preliminary numerical experiments with DM, guided analysis, edited manuscript;
	Y.G.K.: guided the DM, ADM and Y-shaped autoencoder frameworks, edited manuscript;
	A.M.: conceived the idea, initiated project, supervised L.F.K. \& J.M.M.F., wrote initial draft.
	
	\section*{Acknowledgments}
	This work was performed as a part of project B4 of the CRC~985 ``Functional Microgels and Microgel Systems'' funded by Deutsche Forschungsgemeinschaft (DFG). 
	EDK was funded by the Luxembourg National Research Fund (FNR), grant reference 16758846. For the purpose of open access, the authors have applied a Creative Commons Attribution 4.0 International (CC BY 4.0) license to any Author Accepted Manuscript version arising from this submission. 
	The work of YGK was partially supported by the US Air Force Office of Scientific Research. 
	The authors thank J\"orn Viell for scientific discussions and feedback on the manuscript and Andrij Pich for useful discussions on future tasks and impact of this work.
	
	\section*{Data Availability and Reproducibility Statement}
	An invention disclosure is submitted and a patent application is planed for the method.
	The code underlying this contribution is published in a GitLab repository~\citep{Gitlab_DMAPs.2023}.
	The underlying data set is published via RWTH publications~\citep{dataKaven.2023}, including Raman spectra in un-treated and pre-treated form and the according DLS size predictions.
	Data from the manuscript's figures are tabulated in Supplementary Materials.
	
	\newpage
	
	\bibliographystyle{ama.bst}
	\bibliography{bibliographyAIChE.bib}

\begin{thebibliography}{10}

\bibitem{Chew.2010}
Chew W, Sharratt P. Trends in process analytical technology  {\it Analytical
  Methods. } 2010;2:1412.

\bibitem{Beer.2011}
Beer T, Burggraeve A, Fonteyne M, Saerens L, Remon JP, Vervaet C. {Near
  infrared and Raman spectroscopy for the in-process monitoring of
  pharmaceutical production processes}  {\it International journal of
  pharmaceutics. } 2011;417:32--47.

\bibitem{Pomerantsev.2012}
Pomerantsev AL, Rodionova OY. Process analytical technology: a critical view of
  the chemometricians  {\it Journal of Chemometrics. } 2012;26:299--310.

\bibitem{Simon.2015}
Simon LL, Pataki H, Marosi G, et al. {Assessment of Recent Process Analytical
  Technology (PAT) Trends: A Multiauthor Review}  {\it Organic Process Research
  {\&} Development. } 2015;19:3--62.

\bibitem{Beer.1852}
Beer A. Bestimmung der {A}bsorption des rothen {L}ichts in farbigen
  {F}l{\"u}ssigkeiten  {\it Annalen der Physik und Chemie. } 1852;162:78--88.

\bibitem{Marini.2008}
Marini F, Bucci R, Magr{\`i} AL, Magr{\`i} AD. Artificial neural networks in
  chemometrics: History, examples and perspectives  {\it Microchemical Journal.
  } 2008;88:178--185.

\bibitem{Garrido.2008}
Garrido M, Rius FX, Larrechi MS. Multivariate curve resolution-alternating
  least squares ({MCR-ALS}) applied to spectroscopic data from monitoring
  chemical reactions processes  {\it Analytical and bioanalytical chemistry. }
  2008;390:2059--2066.

\bibitem{Alsmeyer2004}
Alsmeyer F, Ko{\ss} H-J, Marquardt W. {Indirect spectral hard modeling for the
  analysis of reactive and interacting mixtures}  {\it {Applied Spectroscopy}.
  } 2004;58:975--985.

\bibitem{Keidel.2018}
Keidel R, Ghavami A, Lugo DM, et al. Time-resolved structural evolution during
  the collapse of responsive hydrogels: The microgel-to-particle transition
  {\it Science advances. } 2018;4:eaao7086.

\bibitem{Bohren.1998}
Bohren Craig~F., Huffman Donald~R.. {\it Absorption and Scattering of Light by
  Small Particles}.
\newblock Wiley 1998.

\bibitem{Reis.2003}
Reis MM, Ara{\'u}jo PHH, Sayer C, Giudici R. Evidences of correlation between
  polymer particle size and {R}aman scattering  {\it Polymer. }
  2003;44:6123--6128.

\bibitem{vandenBrink.2002}
{van den Brink} M, Pepers M, {van Herk} AM. Raman spectroscopy of polymer
  latexes  {\it Journal of Raman Spectroscopy. } 2002;33:264--272.

\bibitem{Ito.2002}
Ito K, Kato T, Ona T. Non-destructive method for the quantification of the
  average particle diameter of latex as water-based emulsions by near-infrared
  {F}ourier transform {R}aman spectroscopy  {\it Journal of Raman Spectroscopy.
  } 2002;33:466--470.

\bibitem{Houben.2015}
Houben C, Nurumbetov G, Haddleton D, Lapkin AA. Feasibility of the Simultaneous
  Determination of Monomer Concentrations and Particle Size in Emulsion
  Polymerization Using in Situ {R}aman Spectroscopy  {\it Industrial {\&}
  engineering chemistry research. } 2015;54:12867--12876.

\bibitem{Ambrogi.2017}
Ambrogi PMN, Colm{\'a}n MME, Giudici R. Miniemulsion Polymerization Monitoring
  Using Off-Line {R}aman Spectroscopy and In-Line {NIR} Spectroscopy  {\it
  Macromolecular Reaction Engineering. } 2017;11:1600013.

\bibitem{MeyerKirschner.2018}
Meyer-Kirschner J, Mitsos A, Viell J. Polymer particle sizing from {R}aman
  spectra by regression of hard model parameters  {\it Journal of Raman
  Spectroscopy. } 2018;49:1402--1411.

\bibitem{r19}
Coifman RR, Lafon S. Diffusion maps  {\it {Applied and Computational Harmonic
  Analysis}. } 2006;21:5--30.

\bibitem{r20}
Nadler B, Lafon S, Coifman RR, Kevrekidis IG. Diffusion maps, spectral
  clustering and reaction coordinates of dynamical systems  {\it Applied and
  Computational Harmonic Analysis. } 2006;21:113--127.

\bibitem{r21}
Coifman RR, Kevrekidis IG, Lafon S, Maggioni M, Nadler B. Diffusion maps,
  reduction coordinates, and low dimensional representation of stochastic
  systems  {\it Multiscale Modeling \& Simulation. } 2008;7:842--864.

\bibitem{lederman2018learning}
Lederman RR, Talmon R. Learning the geometry of common latent variables using
  alternating-diffusion  {\it {Applied and Computational Harmonic Analysis}. }
  2018;44:509--536.

\bibitem{katz2019alternating}
Katz O, Talmon R, Lo~Y-L, Wu~H-T. Alternating diffusion maps for multimodal
  data fusion  {\it Information Fusion. } 2019;45:346-360.

\bibitem{dietrich2022spectral}
Dietrich F, Yair O, Mulayoff R, Talmon R, Kevrekidis IG. Spectral discovery of
  jointly smooth features for multimodal data  {\it SIAM Journal on Mathematics
  of Data Science. } 2022;4:410--430.

\bibitem{yshaped2022}
Evangelou N, Wichrowski NJ, Kevrekidis GA, et al. {On the parameter
  combinations that matter and on those that do not: data-driven studies of
  parameter (non)identifiability}  {\it PNAS Nexus. } 2022;1:pgac154.

\bibitem{r40}
Chiavazzo E, Gear CW, Dsilva CJ, Rabin N, Kevrekidis IG. Reduced models in
  chemical kinetics via nonlinear data-mining  {\it Processes. }
  2014;2:112--140.

\bibitem{KORONAKI2023108357}
Koronaki E, Evangelou N, Psarellis YM, Boudouvis AG, Kevrekidis IG. From
  partial data to out-of-sample parameter and observation estimation with
  diffusion maps and geometric harmonics  {\it Computers \& Chemical
  Engineering. } 2023;178:108357.

\bibitem{nystrom1929praktische}
Nystr{\"o}m EJ. {\it {\"U}ber die praktische {A}ufl{\"o}sung von linearen
  {I}ntegralgleichungen mit {A}nwendungen auf {R}andwertaufgaben der
  {P}otentialtheorie}.
\newblock Akademische Buchhandlung 1929.

\bibitem{fowlkes2001efficient}
Fowlkes C, Belongie S, Malik J. Efficient spatiotemporal grouping using the
  {N}ystrom method  in {\it Proceedings of the 2001 IEEE Computer Society
  Conference on Computer Vision and Pattern Recognition. CVPR 2001};1:I--IIEEE
  2001.

\bibitem{evangelou2022double}
Evangelou N, Dietrich F, Chiavazzo E, Lehmberg D, Meila M, Kevrekidis IG.
  Double Diffusion Maps and their Latent Harmonics for Scientific Computations
  in Latent Space  {\it arXiv preprint arXiv:2204.12536. } 2022.

\bibitem{Kaven.2023}
Kaven LF, Schweidtmann AM, Keil J, Israel J, Wolter N, Mitsos A. {Data-driven
  Product-Process Optimization of N-isopropylacrylamide Microgel Synthesis in
  Flow}  {\it arXiv preprint arXiv:2308.16724. } 2023.

\bibitem{Kaven.2021}
Kaven LF, Wolff HJM, Wille L, Wessling M, Mitsos A, Viell J. {In-line
  Monitoring of Microgel Synthesis: Flow versus Batch Reactor}  {\it Organic
  Process Research {\&} Development. } 2021;25:2039--2051.

\bibitem{Kather.2018}
Kather M, Ritter F, Pich A. {Surfactant-free synthesis of extremely small
  stimuli-responsive colloidal gels using a confined impinging jet reactor}
  {\it Chemical Engineering Journal. } 2018;344:375--379.

\bibitem{Wolff.2018}
Wolff HJM, Kather M, Breisig H, Richtering W, Pich A, Wessling M. {From Batch
  to Continuous Precipitation Polymerization of Thermoresponsive Microgels}
  {\it ACS applied materials {\&} interfaces. } 2018;10:24799--24806.

\bibitem{Janssen.2019}
Janssen FAL, Kather M, Ksiazkiewicz A, Pich A, Mitsos A. {Synthesis of
  Poly(N-vinylcaprolactam)-Based Microgels by Precipitation Polymerization:
  Pseudo-Bulk Model for Particle Growth and Size Distribution}  {\it ACS omega.
  } 2019;4:13795--13807.

\bibitem{dataKaven.2023}
Kaven L, Mitsos A. {Dataset to: Nonlinear Manifold Learning Determines Microgel
  Size from Raman Spectroscopy}   2023.
\newblock doi:\url{10.18154/RWTH-2023-05604 }.

\bibitem{dataKaven.2021}
Kaven LF, Wolff HJ, Wille L, Wessling M, Mitsos A, Viell J. {Dataset to:
  In-line Monitoring of Microgel Synthesis: Flow versus Batch Reactor}   2021.
\newblock doi:\url{10.18154/RWTH-2021-09666}.

\bibitem{psarellis2022data}
Psarellis YM, Lee S, Bhattacharjee T, Datta SS, Bello-Rivas JM, Kevrekidis IG.
  Data-driven discovery of chemotactic migration of bacteria via machine
  learning  {\it arXiv preprint arXiv:2208.11853. } 2022.

\bibitem{coifman2006geometric}
Coifman Ronald~R, Lafon St{\'e}phane. Geometric harmonics: a novel tool for
  multiscale out-of-sample extension of empirical functions  {\it {Applied and
  Computational Harmonic Analysis}. } 2006;21:31--52.

\bibitem{r25}
Dsilva CJ, Talmon R, Coifman RR, Kevrekidis IG. Parsimonious representation of
  nonlinear dynamical systems through manifold learning: A chemotaxis case
  study  {\it {Applied and Computational Harmonic Analysis}. }
  2018;44:759--773.

\bibitem{TALMON2019848}
Talmon R, Wu~H-T. Latent common manifold learning with alternating diffusion:
  Analysis and applications  {\it {Applied and Computational Harmonic
  Analysis}. } 2019;47:848-892.

\bibitem{Gitlab_DMAPs.2023}
Koronaki E, Kaven LF, Faust JMM, Kevrekidis IG, Mitsos A. {Code to: Nonlinear
  Manifold Learning Determines Microgel Size from Raman Spectroscopy}   2023.
\newblock https://gitlab.com/eleni.koronaki/mlforpolymersizeraman.git.

\bibitem{Malvern.2006}
Ltd. Malvern~Instruments. Zetasizer Nano technical note MRK728-01 - The
  accuracy and precision expected from dynamic light scattering measurements
  2006.
\newblock
  \url{https://kdsi.ru/upload/iblock/357/be4ecfa4fcce215f870e4acb8eb229d6.pdf}.

\end{thebibliography}
	
	
	\newpage
	\listoffigures
	
	\newpage
	
	\section*{Figures}
	
	\newpage
	
	\newpage
	
	\newpage
	
	\newpage
	
	\newpage
	
	\newpage
	
	\newpage
	
	\newpage
	
	\newpage
	
	\newpage
	
	\newpage

	\begin{figure}[H]
		\centering
		\begin{subfigure}{0.45\textwidth}
			
			\includegraphics[width=\textwidth]{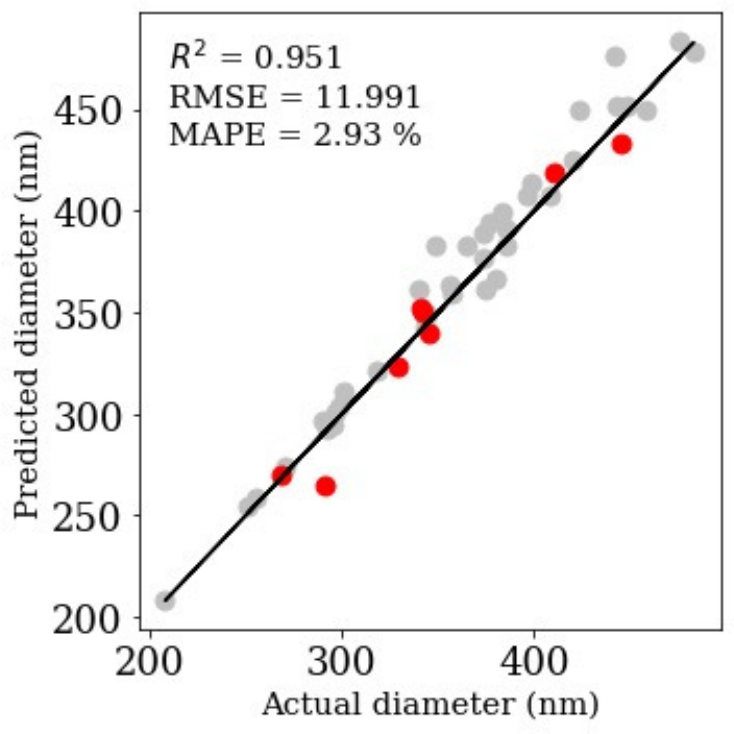}
			\centering
			\caption{}
			\label{fig:yshaped1}
		\end{subfigure}
		\vfill
		\begin{subfigure}{0.45\textwidth}
			\includegraphics[width=\textwidth]{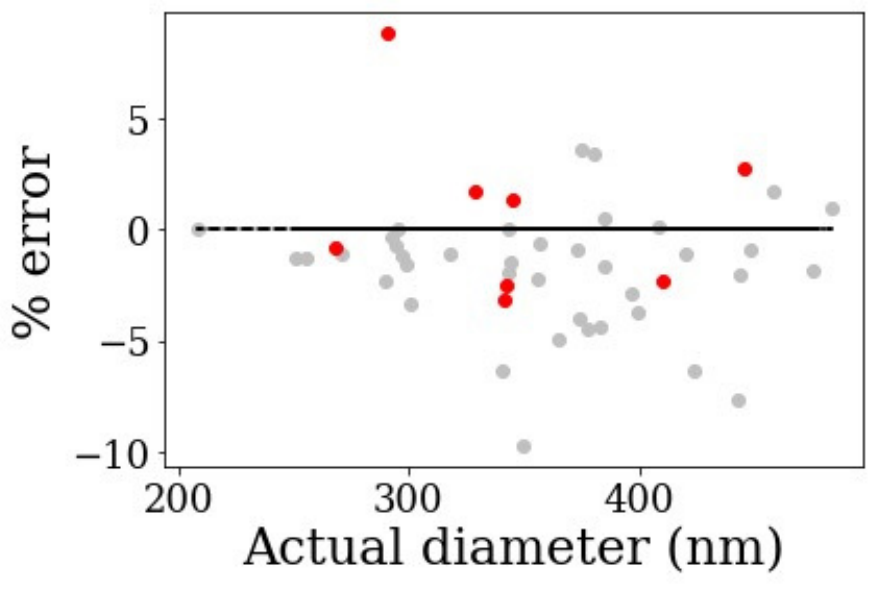}
			\centering
			\caption{}
		\end{subfigure}
		\caption{Prediction from Y-shaped autoencoder architecture: (a) Actual versus predicted polymer size, and (b) $\%$ error for each one of the data points in the test set.
			Red points correspond to the test set, and gray points correspond to the training set values.
			The reported accuracy metrics (R\textsuperscript{2}, RMSE, and MAPE) correspond to the test data.}
		\label{fig:yshaped}
	\end{figure}

	\newpage

	\section*{Tables}
	
	\begin{table}[htbp]
		\centering
		\caption{Overview of state-of-the-art approaches and our proposal (last row) to predict polymer sizes from Raman spectra. 
			The number of data points refers to the amount of samples measured via Raman spectroscopy.}
		\begin{tabular}{p{1.93em}cp{8.645em}ccc}
			\hline
			\multicolumn{1}{c}{Ref.} & Method & \multicolumn{1}{c}{Polymer system} & \multicolumn{1}{p{3.5em}}{Size\newline{}[nm]} & \multicolumn{1}{l}{\# points} & \multicolumn{1}{p{4.5em}}{Spectrum\newline{}[\SI{}{\per\centi\meter}]} \bigstrut\\
			\hline
			\multicolumn{1}{c}{\citep{Ito.2002}} & PLS   & styrene, butadiene, \newline{}methyl methacrylate, acrylonitrile & 80-200 & 47    & 100-4000 \bigstrut\\
			\hline
			\multicolumn{1}{c}{\citep{vandenBrink.2002} } & Focal depth & styrene & 42-210 & 6     & 3200-3800 \bigstrut\\
			\hline
			\multicolumn{1}{c}{\citep{Reis.2003}} & PLS   & styrene, acrylic acid & 55-150 & 23    & 400-4000 \bigstrut\\
			\hline
			\multicolumn{1}{c}{\citep{Houben.2015}} & PLS   & styrene, butyl acrylate, acrylic acid, methyl acrylate & 20-200 & N.S.  & \multicolumn{1}{p{4.5em}}{150-400, \newline{}150-1800} \bigstrut\\
			\hline
			\multicolumn{1}{c}{\citep{Ambrogi.2017} } & PLS   & styrene & 50-400 & 40    & \multicolumn{1}{p{4.5em}}{150-400,\newline{}0-4000} \bigstrut\\
			\hline
			\multicolumn{1}{c}{\citep{MeyerKirschner.2018}} & IHM+PLS & styrene & 23-60 & 21    & 1020-2000 \bigstrut\\
			\hline
			\hline
			\multicolumn{2}{c}{Nonlinear} & N-isopropylacryl- & 208-483 & 47    & 100-3425 \bigstrut\\
			\multicolumn{2}{c}{manifold} & amide & & & \\
			\multicolumn{2}{c}{learning} & & & & \\
			\hline
		\end{tabular}%
		\label{tab:StateOfArt}%
	\end{table}%
	
	\newpage
	\begin{table}[htbp]
		\centering
		\caption{Performance of PLS regression on Raman spectra with different pretreatment methods. 
			The values for RMSE are given in \SI{}{\nano\meter} and for MAPE in~\%.
			The intensity level of the gray shading visualizes the quality of the prediction model compared to the remaining models regarding each prediction performance metric: A lower gray shade corresponds to a better performance.}
		\makebox[\textwidth]{
			
			\begin{tabular}{|>{\hspace{0pt}}m{0.219\linewidth}|>{\hspace{0pt}}m{0.096\linewidth}|>{\hspace{0pt}}S[table-format=1.3,table-column-width=0.075\linewidth]>{\hspace{0pt}}S[table-format=2.2,table-column-width=0.075\linewidth]>{\hspace{0pt}}S[table-format=2.4,table-column-width=0.106\linewidth]|>{\hspace{0pt}}S[table-format=+1.3,table-column-width=0.083\linewidth]>{\hspace{0pt}}S[table-format=2.2,table-column-width=0.079\linewidth]>{\RaggedLeft\hspace{0pt}}m{0.096\linewidth}|>{\Centering\hspace{0pt}}m{0.102\linewidth}|} 
				\hline
				\multirow{2}{0.219\linewidth}{\hspace{0pt}Pretreatment} & Spectral & \multicolumn{3}{>{\Centering\hspace{0pt}}m{0.256\linewidth}|}{Training}                                                                                                                                         & \multicolumn{3}{>{\Centering\hspace{0pt}}m{0.258\linewidth}|}{Testing}                                                                                                                                          & \# latent  \\
				& region   & \multicolumn{1}{>{\Centering\hspace{0pt}}m{0.075\linewidth}}{R\textsuperscript{2}} & \multicolumn{1}{>{\Centering\hspace{0pt}}m{0.075\linewidth}}{RMSE} & \multicolumn{1}{>{\Centering\hspace{0pt}}m{0.106\linewidth}|}{MAPE} & \multicolumn{1}{>{\Centering\hspace{0pt}}m{0.083\linewidth}}{R\textsuperscript{2}} & \multicolumn{1}{>{\Centering\hspace{0pt}}m{0.079\linewidth}}{RMSE} & \multicolumn{1}{>{\Centering\hspace{0pt}}m{0.096\linewidth}|}{MAPE} & variables  \\ 
				\hline
				Linear fit                                              & FP       & {\cellcolor[rgb]{0.929,0.929,0.929}}0.961                            & {\cellcolor[rgb]{0.91,0.91,0.91}}12.664                             & {\cellcolor[rgb]{0.898,0.898,0.898}}2.882                          & {\cellcolor[rgb]{0.804,0.804,0.804}}0.428                            & {\cellcolor[rgb]{0.769,0.769,0.769}}40.865                          & {\cellcolor[rgb]{0.882,0.882,0.882}}8.639                         & 7          \\
				Linear fit                                              & Global   & {\cellcolor[rgb]{0.561,0.561,0.561}}0.284                            & {\cellcolor[rgb]{0.675,0.675,0.675}}54.449                          & {\cellcolor[rgb]{0.537,0.537,0.537}}13.079                          & {\cellcolor[rgb]{0.871,0.871,0.871}}0.573                            & {\cellcolor[rgb]{0.847,0.847,0.847}}35.288                         & {\cellcolor[rgb]{0.894,0.894,0.894}}8.419                         & 2          \\
				MinMax, linear fit                                      & FP       & {\cellcolor[rgb]{0.875,0.875,0.875}}0.860                             & {\cellcolor[rgb]{0.847,0.847,0.847}}24.087                          & {\cellcolor[rgb]{0.82,0.82,0.82}}5.106                             & {\cellcolor[rgb]{0.651,0.651,0.651}}0.106                            & {\cellcolor[rgb]{0.624,0.624,0.624}}51.080                          & {\cellcolor[rgb]{0.651,0.651,0.651}}12.935                         & 7          \\
				MinMax, linear fit                                      & Global   & {\cellcolor[rgb]{0.631,0.631,0.631}}0.416                            & {\cellcolor[rgb]{0.706,0.706,0.706}}49.157                          & {\cellcolor[rgb]{0.596,0.596,0.596}}11.369                          & {\cellcolor[rgb]{0.929,0.929,0.929}}0.696                            & {\cellcolor[rgb]{0.925,0.925,0.925}}29.795                          & {\cellcolor[rgb]{0.949,0.949,0.949}}7.389                         & 3          \\
				MinMax, rubber band                                     & FP       & {\cellcolor[rgb]{0.875,0.875,0.875}}0.861                            & {\cellcolor[rgb]{0.847,0.847,0.847}}23.960                          & {\cellcolor[rgb]{0.812,0.812,0.812}}5.345                          & {\cellcolor[rgb]{0.502,0.502,0.502}}-0.206                           & {\cellcolor[rgb]{0.502,0.502,0.502}}59.322                          & {\cellcolor[rgb]{0.502,0.502,0.502}}15.700                         & 6          \\
				MinMax, rubber band                                     & Global   & {\cellcolor[rgb]{0.635,0.635,0.635}}0.425                            & {\cellcolor[rgb]{0.706,0.706,0.706}}48.794                          & {\cellcolor[rgb]{0.6,0.6,0.6}}11.232                                & {\cellcolor[rgb]{0.929,0.929,0.929}}0.697                            & {\cellcolor[rgb]{0.925,0.925,0.925}}29.740                          & {\cellcolor[rgb]{0.945,0.945,0.945}}7.496                         & 3          \\ 
				\hline
				Raw                                                     & FP       & {\cellcolor[rgb]{0.835,0.835,0.835}}0.790                             & {\cellcolor[rgb]{0.816,0.816,0.816}}29.463                          & {\cellcolor[rgb]{0.761,0.761,0.761}}6.750                          & {\cellcolor[rgb]{0.949,0.949,0.949}}0.730                             & {\cellcolor[rgb]{0.949,0.949,0.949}}28.038                          & {\cellcolor[rgb]{0.925,0.925,0.925}}7.887                         & 5          \\ 
				\hline
				Raw                                                     & Global   & {\cellcolor[rgb]{0.918,0.918,0.918}}0.942                            & {\cellcolor[rgb]{0.894,0.894,0.894}}15.473                          & {\cellcolor[rgb]{0.867,0.867,0.867}}3.799                          & {\cellcolor[rgb]{0.902,0.902,0.902}}0.633                            & {\cellcolor[rgb]{0.886,0.886,0.886}}32.701                           & {\cellcolor[rgb]{0.922,0.922,0.922}}7.953                         & 8          \\ 
				\hline
				Rubber band                                             & FP       & {\cellcolor[rgb]{0.949,0.949,0.949}}0.993                            & {\cellcolor[rgb]{0.949,0.949,0.949}}5.445                           & {\cellcolor[rgb]{0.949,0.949,0.949}}1.359                          & {\cellcolor[rgb]{0.58,0.58,0.58}}-0.041                              & {\cellcolor[rgb]{0.565,0.565,0.565}}55.109                          & {\cellcolor[rgb]{0.576,0.576,0.576}}14.329                         & 9          \\
				Rubber band                                             & Global   & {\cellcolor[rgb]{0.682,0.682,0.682}}0.509                            & {\cellcolor[rgb]{0.729,0.729,0.729}}45.073                          & {\cellcolor[rgb]{0.616,0.616,0.616}}10.800                            & {\cellcolor[rgb]{0.914,0.914,0.914}}0.662                            & {\cellcolor[rgb]{0.902,0.902,0.902}}31.401                           & {\cellcolor[rgb]{0.937,0.937,0.937}}7.614                         & 3          \\
				SNV, linear fit                                         & FP       & {\cellcolor[rgb]{0.502,0.502,0.502}}0.181                            & {\cellcolor[rgb]{0.655,0.655,0.655}}58.212                          & {\cellcolor[rgb]{0.502,0.502,0.502}}14.004                          & {\cellcolor[rgb]{0.894,0.894,0.894}}0.616                            & {\cellcolor[rgb]{0.875,0.875,0.875}}33.455                          & {\cellcolor[rgb]{0.902,0.902,0.902}}8.286                         & 1          \\
				SNV, linear fit                                         & Global   & {\cellcolor[rgb]{0.918,0.918,0.918}}0.937                            & {\cellcolor[rgb]{0.89,0.89,0.89}}16.184                             & {\cellcolor[rgb]{0.859,0.859,0.859}}3.944                          & {\cellcolor[rgb]{0.757,0.757,0.757}}0.329                            & {\cellcolor[rgb]{0.718,0.718,0.718}}44.230                          & {\cellcolor[rgb]{0.824,0.824,0.824}}9.777                         & 6          \\
				SNV, rubber band                                        & FP       & {\cellcolor[rgb]{0.502,0.502,0.502}}0.175                            & {\cellcolor[rgb]{0.651,0.651,0.651}}58.431                          & {\cellcolor[rgb]{0.506,0.506,0.506}}13.992                          & {\cellcolor[rgb]{0.894,0.894,0.894}}0.620                             & {\cellcolor[rgb]{0.875,0.875,0.875}}33.300                           & {\cellcolor[rgb]{0.902,0.902,0.902}}8.336                         & 1          \\
				SNV, rubber band                                        & Global   & {\cellcolor[rgb]{0.918,0.918,0.918}}0.937                            & {\cellcolor[rgb]{0.89,0.89,0.89}}16.173                    & {\cellcolor[rgb]{0.859,0.859,0.859}}3.942                          & {\cellcolor[rgb]{0.761,0.761,0.761}}0.342                            & {\cellcolor[rgb]{0.725,0.725,0.725}}43.821                          & {\cellcolor[rgb]{0.827,0.827,0.827}}9.708                         & 6          \\
				\hline
			\end{tabular}
		}
		\label{tab:results_PLS}%
	\end{table}%
	
	\newpage
	
	\begin{table}
		\centering
		
		\caption{Performance of PLS regression on IHM parameters from Raman spectra with different pretreatment methods.
			The values for RMSE are given in \SI{}{\nano\meter} and for MAPE in~\%.
			The intensity level of the gray shading visualizes the quality of the prediction model compared to the remaining models regarding each prediction performance metric: A lower gray shade corresponds to a better performance.}
		\makebox[\textwidth]{ 
			
			\begin{tabular}{|>{\hspace{0pt}}m{0.229\linewidth}|>{\hspace{0pt}}m{0.094\linewidth}!{\color{black}\vrule}>{\hspace{0pt}}S[table-format=1.3,table-column-width=0.073\linewidth]>{\hspace{0pt}}S[table-format=2.3,table-column-width=0.087\linewidth]>{\hspace{0pt}}S[table-format=2.3,table-column-width=0.087\linewidth]!{\color{black}\vrule}>{\hspace{0pt}}S[table-format=+1.3,table-column-width=0.083\linewidth]>{\hspace{0pt}}S[table-format=2.3,table-column-width=0.087\linewidth]>{\hspace{0pt}}S[table-format=2.3,table-column-width=0.087\linewidth]!{\color{black}\vrule}>{\centering\arraybackslash\hspace{0pt}}m{0.106\linewidth}|} 
				\hline
				\multirow{2}{0.229\linewidth}{\hspace{0pt}Pretreatment} & \multicolumn{1}{>{\hspace{0pt}}m{0.094\linewidth}|}{Fitting} & \multicolumn{3}{>{\centering\hspace{0pt}}m{0.246\linewidth}|}{Training}                                                                                                                                         & \multicolumn{3}{>{\centering\hspace{0pt}}m{0.257\linewidth}|}{Testing}                                                                                                                                          & \# latent  \\
				& \multicolumn{1}{>{\hspace{0pt}}m{0.094\linewidth}|}{mode}    & \multicolumn{1}{>{\centering\hspace{0pt}}m{0.073\linewidth}}{R\textsuperscript{2}} & \multicolumn{1}{>{\centering\hspace{0pt}}m{0.087\linewidth}}{RMSE} & \multicolumn{1}{>{\centering\hspace{0pt}}m{0.087\linewidth}|}{MAPE} & \multicolumn{1}{>{\centering\hspace{0pt}}m{0.083\linewidth}}{R\textsuperscript{2}} & \multicolumn{1}{>{\centering\hspace{0pt}}m{0.087\linewidth}}{RMSE} & \multicolumn{1}{>{\centering\hspace{0pt}}m{0.087\linewidth}|}{MAPE} & variables  \\ 
				\hline
				Linear fit                                              & High                                                         & {\cellcolor[rgb]{0.827,0.827,0.827}}0.577                            & {\cellcolor[rgb]{0.776,0.776,0.776}}41.834                         & {\cellcolor[rgb]{0.816,0.816,0.816}}10.142                          & {\cellcolor[rgb]{0.812,0.812,0.812}}0.472                            & {\cellcolor[rgb]{0.78,0.78,0.78}}39.245                            & {\cellcolor[rgb]{0.894,0.894,0.894}}9.014                           & 2          \\
				Linear fit                                              & Medium                                                       & {\cellcolor[rgb]{0.502,0.502,0.502}}0.327                            & {\cellcolor[rgb]{0.502,0.502,0.502}}52.768                         & {\cellcolor[rgb]{0.502,0.502,0.502}}12.145                          & {\cellcolor[rgb]{0.8,0.8,0.8}}0.316                                  & {\cellcolor[rgb]{0.769,0.769,0.769}}44.667                         & {\cellcolor[rgb]{0.8,0.8,0.8}}11.785                                & 2          \\
				Raw                                                     & High                                                         & {\cellcolor[rgb]{0.949,0.949,0.949}}0.839                            & {\cellcolor[rgb]{0.949,0.949,0.949}}25.817                         & {\cellcolor[rgb]{0.949,0.949,0.949}}6.455                           & {\cellcolor[rgb]{0.502,0.502,0.502}}-0.257                           & {\cellcolor[rgb]{0.502,0.502,0.502}}60.565                         & {\cellcolor[rgb]{0.502,0.502,0.502}}14.040                          & 6          \\
				Raw                                                     & Medium                                                       & {\cellcolor[rgb]{0.502,0.502,0.502}}0.327                            & {\cellcolor[rgb]{0.506,0.506,0.506}}52.780                         & {\cellcolor[rgb]{0.506,0.506,0.506}}12.145                          & {\cellcolor[rgb]{0.808,0.808,0.808}}0.329                            & {\cellcolor[rgb]{0.776,0.776,0.776}}44.237                         & {\cellcolor[rgb]{0.808,0.808,0.808}}11.637                          & 2          \\
				Rubber band                                             & High                                                         & {\cellcolor[rgb]{0.776,0.776,0.776}}0.646                            & {\cellcolor[rgb]{0.722,0.722,0.722}}38.255                         & {\cellcolor[rgb]{0.757,0.757,0.757}}8.739                           & {\cellcolor[rgb]{0.835,0.835,0.835}}0.335                            & {\cellcolor[rgb]{0.804,0.804,0.804}}44.043                         & {\cellcolor[rgb]{0.816,0.816,0.816}}10.582                          & 6          \\
				Rubber band                                             & Medium                                                       & {\cellcolor[rgb]{0.592,0.592,0.592}}0.538                            & {\cellcolor[rgb]{0.569,0.569,0.569}}43.745                         & {\cellcolor[rgb]{0.608,0.608,0.608}}9.917                           & {\cellcolor[rgb]{0.643,0.643,0.643}}-0.019                           & {\cellcolor[rgb]{0.62,0.62,0.62}}54.518                            & {\cellcolor[rgb]{0.788,0.788,0.788}}10.765                          & 8          \\ 
				\hline
				MinMax, linear fit                                      & High                                                         & {\cellcolor[rgb]{0.831,0.831,0.831}}0.741                            & {\cellcolor[rgb]{0.78,0.78,0.78}}32.746                            & {\cellcolor[rgb]{0.82,0.82,0.82}}7.265                              & {\cellcolor[rgb]{0.808,0.808,0.808}}0.278                            & {\cellcolor[rgb]{0.776,0.776,0.776}}45.907                         & {\cellcolor[rgb]{0.89,0.89,0.89}}9.243                              & 5          \\ 
				\hline
				MinMax, linear fit                                      & Medium                                                       & {\cellcolor[rgb]{0.686,0.686,0.686}}0.359                            & {\cellcolor[rgb]{0.639,0.639,0.639}}51.491                         & {\cellcolor[rgb]{0.663,0.663,0.663}}11.449                          & {\cellcolor[rgb]{0.533,0.533,0.533}}0.330                            & {\cellcolor[rgb]{0.529,0.529,0.529}}44.198                         & {\cellcolor[rgb]{0.541,0.541,0.541}}11.980                          & 2          \\
				MinMax, rubber band                                     & High                                                         & {\cellcolor[rgb]{0.796,0.796,0.796}}0.680                            & {\cellcolor[rgb]{0.741,0.741,0.741}}36.396                         & {\cellcolor[rgb]{0.792,0.792,0.792}}7.914                           & {\cellcolor[rgb]{0.831,0.831,0.831}}0.323                            & {\cellcolor[rgb]{0.8,0.8,0.8}}44.433                               & {\cellcolor[rgb]{0.804,0.804,0.804}}11.075                          & 6          \\
				MinMax, rubber band                                     & Medium                                                       & {\cellcolor[rgb]{0.608,0.608,0.608}}0.508                            & {\cellcolor[rgb]{0.58,0.58,0.58}}45.108                            & {\cellcolor[rgb]{0.627,0.627,0.627}}9.986                           & {\cellcolor[rgb]{0.631,0.631,0.631}}-0.153                           & {\cellcolor[rgb]{0.608,0.608,0.608}}57.985                         & {\cellcolor[rgb]{0.808,0.808,0.808}}11.675                          & 8          \\
				SNV, linear fit                                         & High                                                         & {\cellcolor[rgb]{0.835,0.835,0.835}}0.681                            & {\cellcolor[rgb]{0.784,0.784,0.784}}36.312                         & {\cellcolor[rgb]{0.816,0.816,0.816}}8.009                           & {\cellcolor[rgb]{0.792,0.792,0.792}}0.287                            & {\cellcolor[rgb]{0.757,0.757,0.757}}45.599                         & {\cellcolor[rgb]{0.859,0.859,0.859}}8.683                           & 4          \\
				SNV, linear fit                                         & Medium                                                       & {\cellcolor[rgb]{0.682,0.682,0.682}}0.366                            & {\cellcolor[rgb]{0.635,0.635,0.635}}51.220                         & {\cellcolor[rgb]{0.659,0.659,0.659}}11.413                          & {\cellcolor[rgb]{0.529,0.529,0.529}}0.283                            & {\cellcolor[rgb]{0.525,0.525,0.525}}45.723                         & {\cellcolor[rgb]{0.541,0.541,0.541}}12.380                          & 2          \\ 
				\hline
				SNV, rubber band                                        & High                                                         & {\cellcolor[rgb]{0.733,0.733,0.733}}0.585                            & {\cellcolor[rgb]{0.682,0.682,0.682}}41.465                         & {\cellcolor[rgb]{0.714,0.714,0.714}}9.559                           & {\cellcolor[rgb]{0.949,0.949,0.949}}0.611                            & {\cellcolor[rgb]{0.949,0.949,0.949}}33.691                         & {\cellcolor[rgb]{0.949,0.949,0.949}}8.258                           & 4          \\ 
				\hline
				SNV, rubber band                                        & Medium                                                       & {\cellcolor[rgb]{0.608,0.608,0.608}}0.511                            & {\cellcolor[rgb]{0.58,0.58,0.58}}44.967                            & {\cellcolor[rgb]{0.624,0.624,0.624}}10.015                          & {\cellcolor[rgb]{0.627,0.627,0.627}}-0.104                           & {\cellcolor[rgb]{0.608,0.608,0.608}}56.741                         & {\cellcolor[rgb]{0.804,0.804,0.804}}11.607                          & 8          \\
				\hline
			\end{tabular}
		}
		\label{tab:results_IHM_PLS}
	\end{table}
	
	\newpage
	
	\begin{table}[]
		\caption{Testing performance of all considered prediction methods within this work.
			The values for RMSE are given in \SI{}{\nano\meter} and for MAPE in \%.}
		\label{tab:comparison_summary}
		\resizebox{\textwidth}{!}{%
			
			\begin{tabular}{|l|l|c|c|c|c|}
				\hline
				Cluster & Method & \multicolumn{1}{l|}{R\textsuperscript{2}} & \multicolumn{1}{l|}{RMSE} & \multicolumn{1}{l|}{MAPE} & \# latent  \\ 
				&&&&&variables \\ \hline
				\multirow{2}{*}{State-of-the-art} & \begin{tabular}[c]{@{}l@{}}Best configuration based on PLS\\ regression directly to Raman spectra\end{tabular} & 0.633 & 32.701 & 7.953 & 8\\ \cline{2-6} 
				& \begin{tabular}[c]{@{}l@{}}Best configuration based on PLS\\ regression on IHM parameters\end{tabular} & 0.611 & 33.691 & 8.258 & 4\\ \hline
				\multirow{4}{*}{\begin{tabular}[c]{@{}l@{}}Alternating Diffusion\\ Maps\end{tabular}} & \begin{tabular}[c]{@{}l@{}}Neural network based on \\ DoubleDMAPs-predicted AltDMAPs\end{tabular} & 0.584 & 68.162 & 8.101 & 2(+4)\\ \cline{2-6} 
				& \begin{tabular}[c]{@{}l@{}}Neural network based on \\ XGBOOST-predicted AltDMAPs\end{tabular} & 0.330 & 70.130 & 10.058 & 2 (+4) \\ \cline{2-6} 
				& \begin{tabular}[c]{@{}l@{}}XGBOOST based on \\ DoubleDMAPs-predicted AltDMAPs\end{tabular} & 0.345 & 43.717 & 10.772 & 2 (+4) \\ \cline{2-6} 
				& \begin{tabular}[c]{@{}l@{}}XGBOOST based on \\ XGBOOST-predicted AltDMAPs\end{tabular} & 0.525 & 37.222 & 8.481 & 2 (+4) \\ \hline
				\multirow{3}{*}{\begin{tabular}[c]{@{}l@{}}Prediction directly from \\ DMAP coordinates\end{tabular}} & Neural network & 0.538 & 78.880 & 7.895 & 6\\ \cline{2-6} 
				& XGBOOST & 0.600 & 34.172 & 6.348 & 6\\ \cline{2-6} 
				& Y-shaped autoencoder & 0.951 & 11.991 & 2.930 & 1\\ \hline
				
			\end{tabular}%
		}
	\end{table}

\end{document}